\documentclass[sigconf]{acmart}

\usepackage{booktabs} 

\usepackage{helvet}  
\usepackage{url}  
\usepackage{graphicx}
\usepackage{caption}
\usepackage{courier}
\usepackage{multirow}
\usepackage{color}
\usepackage{subcaption}
\usepackage{color}
\usepackage{todonotes}
\usepackage{amssymb}
\usepackage{gensymb}
\usepackage{multirow}
\usepackage[linesnumbered, ruled]{algorithm2e}
\usepackage{amsmath}

\usepackage{helvet}  
\usepackage{courier}  
\usepackage{url}  
\usepackage{graphicx}  
\usepackage{xspace}
\usepackage{amsmath}
\usepackage{xcolor}
\usepackage{multirow}
\usepackage{caption}
\usepackage{subcaption}
\newcommand{\ce}{{JMDI}\xspace}

\copyrightyear{2019}
\acmYear{2019} 
\setcopyright{iw3c2w3}
\acmConference[WWW '19]{Proceedings of the 2019 World Wide Web Conference}{May 13--17, 2019}{San Francisco, CA, USA}
\acmBooktitle{Proceedings of the 2019 World Wide Web Conference (WWW'19), May 13--17, 2019, San Francisco, CA, USA}
\acmPrice{}
\acmDOI{10.1145/3308558.3313621}
\acmISBN{978-1-4503-6674-8/19/05}

\usepackage{balance}

\begin{document}

\title{Joint Modeling of Dense and Incomplete Trajectories for Citywide Traffic Volume Inference}
\author{Xianfeng Tang{$^{1}$}, Boqing Gong{$^{2}$}, Yanwei Yu{$^{3}$}, Huaxiu Yao{$^{1}$}, Yandong Li{$^{4}$}}
\author{Haiyong Xie{$^{5}$}, Xiaoyu Wang{$^{6}$}}
\affiliation{%
  \institution{{\small $^{1}$}The Pennsylvania State University, USA}
}
\affiliation{%
  \institution{{$^{2}$} Tencent AI Lab}
}
\affiliation{%
  \institution{{$^{3}$} Yantai University, China}
}
\affiliation{%
  \institution{{$^{4}$}University of Central Florida, USA}
}

\affiliation{%
  \institution{{$^{5}$}University of Science and Technology of China, China}
}

\affiliation{%
  \institution{{$^{6}$}Intellifusion Inc.}
}
\email{{tangxianfeng,boqinggo,lyndon.leeseu}@outlook.com,haiyong.xie@ieee.org}
\email{{yuyanwei0530, yhx662012, fanghuaxue}@gmail.com}

\renewcommand{\shortauthors}{X. Tang et al.}

\begin{abstract}
Real-time traffic volume inference is key to an intelligent city. It is a challenging task because accurate traffic volumes on the roads can only be measured at certain locations where sensors are installed. Moreover, the traffic evolves over time due to the influences of weather, events, holidays, etc.  Existing solutions to the traffic volume inference problem often rely on dense GPS trajectories, which inevitably fail to account for the vehicles which carry no  GPS devices or have them turned off. Consequently, the results are biased to taxicabs because they are almost always online for GPS tracking. In this paper, we propose a novel framework for the citywide traffic volume inference using both dense GPS trajectories and incomplete trajectories captured by camera surveillance systems. Our approach employs a high-fidelity traffic simulator and deep reinforcement learning to recover full vehicle movements from the incomplete trajectories. In order to jointly model the recovered trajectories and dense GPS trajectories, we construct spatiotemporal graphs and use
multi-view graph embedding to encode the multi-hop correlations between road segments into real-valued vectors. Finally, we infer the citywide traffic volumes by propagating the traffic values of monitored road segments to the unmonitored ones through masked pairwise similarities. Extensive experiments with two big regions in a provincial capital city in China verify the effectiveness of our approach.
\end{abstract}

%
%


\maketitle
\section{Introduction}
With the vast deployment of sensors, such as loop detectors and surveillance cameras, and the rapid development of data storage techniques, an intelligent city can collect and store a large amount of traffic volume data on the daily basis. It is vital to not only monitor ``visible'' traffic volumes in real time but also infer those unseen by the sensors. Information extracted from the traffic volume data may contribute to a wide range of urban applications, such as route selection, traffic control, and urban planning, etc.

However, the sensors are often displaced away from each other to a certain distance, and some of them could fail now and then. As a result, the citywide traffic volume data inevitably have lots of missing values, causing challenges to many downstream applications. In this paper, we aim to develop techniques for the citywide traffic volume inference --- in other words, to estimate the traffic volume  of each and every road segment in every time interval in the city.

The citywide traffic volume inference is a very challenging problem for two major reasons. Firstly, sensors to monitor the traffic are not only sparse but also non-uniform, resulting in relatively low spatial coverage~\cite{li2017citywide}.
It is difficult to obtain any historic traffic data for the currently unmonitored roads.  Secondly, traffic patterns evolve over time and could suddenly change rapidly due to big events. The high dynamism of traffic makes it  difficult to infer the traffic volume data by drawing similarities between different road segments or across various time intervals. 

Existing works on inferring the traffic volume mostly fall into the category of missing (spatiotemporal) data inference. A basic and representative framework is to estimate the missing values by linear interpolation~\cite{hu2016crowdsourcing,shan2013urban,zheng2013time,aslam2012city}. However, those  approaches fail to maintain the global consistency of the inferred results and assume simple structures underlying the historic spatiotemporal data. Another frequently used method is collaborative filtering~\cite{yi2016st}. However, this approach fails for the unmonitored road segments which have no historical information.  

Some recent studies use dense trajectories 
from mobile devices for traffic volume inference. \citeauthor{zhan2017citywide} estimate the citywide traffic volume by combining GPS trajectories with road networks, point-of-interest information, and weather conditions. 
\citeauthor{meng2017city} characterize the similarities between roads  using travel speeds extracted from taxi trajectories.
However, the dense trajectories, no matter extracted from GPS or taxi, are biased representations of the real traffic. After all, taxi is only a small portion of the total vehicles and not all vehicles carry or turn on the GPS devices. 

Partially observed trajectories, which are captured by the static and discrete sensors (e.g., a vehicle tracked by a camera network), are notably under-explored by the existing works. Unlike the dense trajectories (e.g., generated by ``active'' GPS devices), the partially observed trajectories are incomplete because they are obtained from the ``passive'' static sensors (e.g., cameras). As a result, they are available for almost all kinds of vehicles no matter they have GPS devices installed or not.
Compared with dense trajectories, such incomplete trajectories are collected in a much bigger scale everyday. For instance, the surveillance system recognizes  and provides the incomplete trajectories of more than 1.1 million unique vehicles in a provincial capital city in China every single day at the time of this paper written. To the best of our knowledge, the incomplete trajectories have not been utilized before for the traffic volume inference.

Of course, it is more involved to model the spatiotemporal patterns of incomplete trajectories than their dense counterparts. The uncertainties between monitor points have to be taken care of, as well as the noisy measurements at the monitor points. There exist some research tackling the uncertainty problem. For example, \citeauthor{zheng2012reducing} investigate how to reduce the uncertainty in low-sampling-rate trajectories by inferring the possible routes therein. 
\citeauthor{banerjee2014inferring} employ road-network constraints to summarize all probable routes in a holistic manner.
\citeauthor{li2015inferring} use the GPS snippets to help recover the full trajectories.
However, these methods require that every road segment has appeared at least once in the past
in order to guide the recovering procedure of the full trajectories. Besides, their performance significantly degrades when the traffic pattern changes, e.g., due to big events or severe weather. 

To address the aforementioned challenges, we propose a novel framework to \textbf{J}ointly \textbf{M}odel the \textbf{D}ense and \textbf{I}ncomplete trajectories (\ce) for the citywide traffic volume inference. Specifically, we first recover detailed vehicle movements given the incomplete trajectories using a high-fidelity city traffic simulator. We tune the simulator's parameters using deep reinforcement learning so as to automatically identify the proper simulator state which matches the real data. After that, two spatiotemporal graphs are constructed using the recovered trajectories and the dense GPS trajectories, respectively. The graphs are a natural choice for modeling the dynamism of traffic volumes of road segments and over time. Meanwhile, the graphs enable us to conveniently embed the road segments (nodes of the graph) to continuous-value vectors. Finally, we use these embeddings to construct pairwise similarities between the road segments and then propagate the known traffic volumes to the road segments of unknown traffic. 

We conduct extensive experiments on large-scale real-world traffic datasets collected from a provincial capital city in China. The dense trajectories are available for GPS-enabled taxi. For the incomplete trajectories, we extract partially observed information from the surveillance cameras as a result of the plate number identification technique~\cite{duan2005building}. We evaluate the proposed framework \ce against competitive baselines on two selected regions of the city and during different time periods. We also study some ablations of our approach to highlight the effectiveness of main components of \ce. 

We summarize our main contribution as follows.
\begin{itemize}
\item We propose a novel framework, called \ce, to infer citywide traffic volumes using both dense and incomplete traffic trajectories.
\item We propose to use a high-fidelity traffic simulator to recover full vehicle movements from incomplete trajectories. Moreover, in order to efficiently tune the simulator towards the real data, we design a deep reinforcement learning algorithm to automatically set the parameters of the simulator. 
\item We present a joint embedding method to learn meaningful representations of road segments using spatiotemporal graphs. We further use a semi-supervised propagation method to infer citywide traffic volume values.

\item We conduct extensive experiments on large-scale real-world traffic datasets to validate the proposed approach.
\end{itemize}
To the best of our knowledge, this work is the first to jointly model the dense and incomplete trajectories for the citywide traffic volume inference. Since the dense GPS trajectories are biased to taxi, the incomplete trajectories obtained from other sensors show strong complement to their dense counterparts. Jointly, they give rise to better results than either individually.

\section{Related Work}
\subsection{Traffic Volume Inference}
Some studies \cite{hu2016crowdsourcing,shan2013urban,zheng2013time} use linear regression models to infer missing traffic speed or travel time based on taxi trajectories. \citeauthor{aslam2012city}~\cite{aslam2012city} learn a regression model with taxi GPS trajectories to estimate traffic volume. \emph{However, regression methods require a great amount of labeled training data, which is unavailable in our problem setting.}

Another category of prior studies apply principal component analysis (PCA) (e.g., ~\cite{qu2008bpca,qu2009ppca,li2013efficient,asif2016matrix}) or collaborative filtering (CF) (e.g., ~\cite{yi2016st,ruan2017recovering,wang2014travel}) to fill in missing values in spatiotemporal data. PCA-based methods extract traffic patterns from observed data using various PCA techniques, such as Bayesian PCA~\cite{qu2008bpca}, Probabilistic PCA~\cite{qu2009ppca,li2013efficient} and FPCA~\cite{asif2016matrix}. 
CF-based methods recover missing values by decomposing spatiotemporal data into the product of low-rank matrices. 
\citeauthor{yi2016st}~\cite{yi2016st} fill missing values in geo-sensory time series using CF from multiple spatial and temporal perspectives. 
\citeauthor{ruan2017recovering}~\cite{ruan2017recovering} use tensor completion to recover missing values in spatiotemporal sensory data. 
\citeauthor{wang2014travel}~\cite{wang2014travel} propose a tensor factorization method to estimate the missing travel time for drivers on road segments.
\emph{However, both PCA-based and CF-based methods rely on historical data when filling in. They are unable to handle our problem since the traffic volume of unmonitored road segments are totally missing.}

Graph-based Semi-supervised learning (SSL) method~\cite{culp2008graph} has been widely applied for unlabeled data inference, which can be used for inferring missing values. Label propagation~\cite{zhu2003semi}, a classic semi-supervised method, infers unobserved labels by propagating existing labels on an affinity graph. Other SSL based methods \cite{zhou2004learning,argyriou2006combining,yamaguchi2015omni} have been proposed to model the similarities of vertices in the affinity graph. 
\emph{Although Graph-based SSL methods can be applied to our problem, they only consider the similarities between vertices. Therefore, they fail to utilize rich traffic flow information for volume inference.}

Other existing work aim to infer traffic volume values of road segments using loop detector~\cite{kwon2003estimation,wilkie2013flow,meng2017city}, surveillance cameras~\cite{zhan2015lane}, or float car trajectories~\cite{zhan2017citywide,aslam2012city,guhnemann2004monitoring}.
Studies~\cite{kwon2003estimation,wilkie2013flow,zhan2015lane} tackle the volume estimation of a single road segment with loop detectors or surveillance cameras. Thus their methods cannot infer citywide traffic volume.
\citeauthor{zhan2017citywide}~\cite{zhan2017citywide} propose a method to estimate citywide traffic volume using probe taxi trajectories. They estimate travel speeds for volume inference using full taxi trajectories.
Recently, \citeauthor{meng2017city}~\cite{meng2017city} propose ST-SSL that predicts city-wide traffic volume values using loop detector incorporating taxi trajectories. They first build a spatiotemporal affinity graph based on travel speed pattern and spatial correlations extracted from loop detector and taxi trajectories. Then they infer traffic volume using semi-supervised learning method on the spatiotemporal affinity graph.
\emph{However, both \cite{zhan2017citywide} and ST-SSL \cite{meng2017city} require full observation of trajectories, thus they are unable to utilize incomplete counterparts.}

\subsection{Trajectory Recovery}
Several methods (e.g., \cite{wei2012constructing,zheng2012reducing,banerjee2014inferring}) have been proposed to predict complete trajectories from partial observations. 
\citeauthor{zheng2012reducing}~\cite{zheng2012reducing} investigate how to reduce the uncertainty in low-sampling-rate trajectories. This approach requires every road segment appearing in historical data, which is impossible for incomplete trajectories.
\citeauthor{banerjee2014inferring}~\cite{banerjee2014inferring} infer uncertain trajectories from network-constrained partial observations by summarizing all probable routes in a holistic manner. However, their method ignores many important factors such as vehicle interaction and road constraints.
\emph{More important, these methods require historical trajectories as input, which cannot be applied to recovering routes in our problem.}

\section{Notations and Problem Statement}
Let $\{r_1,r_2, \dots, r_m\}$ denote road segments of a road network in a city region. Each road segment is a directed short portion of a road. The traffic volume of a road segment can be represented using the total number of vehicles traversing through it during a fixed time interval. We split the whole time period (e.g, one week) into equal-length continuous time intervals and use  vector $\mathbf{x}_i = (x_{i1}, x_{i2}, \dots, x_{in})$ to denote the traffic volume values of the road segment $r_i$ during the past $n$ continuous time intervals. Note that the traffic volume values are only available at monitored road segments.

Different sources of trajectories  (i.e., incomplete observations from cameras or GPS trackers of taxi) are denotes by $\Delta^{\rm S} = \{\mathbf{\delta}_i\}$, where the superscript $\rm S$ stands for the source $\rm S$. A trajectory $\mathbf{\delta}_i = p_{i,1} \rightarrow p_{i,2} \rightarrow \dots \rightarrow p_{i,s}$  is a list of $s$ points in chronological order, where each point $p = \langle l, t \rangle$ is a location coordinate $l$ (i.e., latitude and longitude) and a time stamp $t$. We denote by $\Delta^{\rm D}$ and $\Delta^{\rm I}$ dense and incomplete trajectories, respectively. Note that points contained in the incomplete trajectories $\Delta^{\rm I}$ are restricted to monitor points --- therefore they form a fixed finite set.

The citywide traffic volume inference problem is defined as follows:
Given the road network, observed traffic volume values $\{x_{it}\}$ at monitored road segments, and multiple sources of trajectories $\Delta^{\rm S}$ ($\rm S \in \rm \{\rm D, I\}$), our goal is to infer the traffic volume values of any unmonitored road segment at any time interval.

\section{Approach}
\begin{figure*}
\begin{center}
\includegraphics[width=0.8\textwidth]{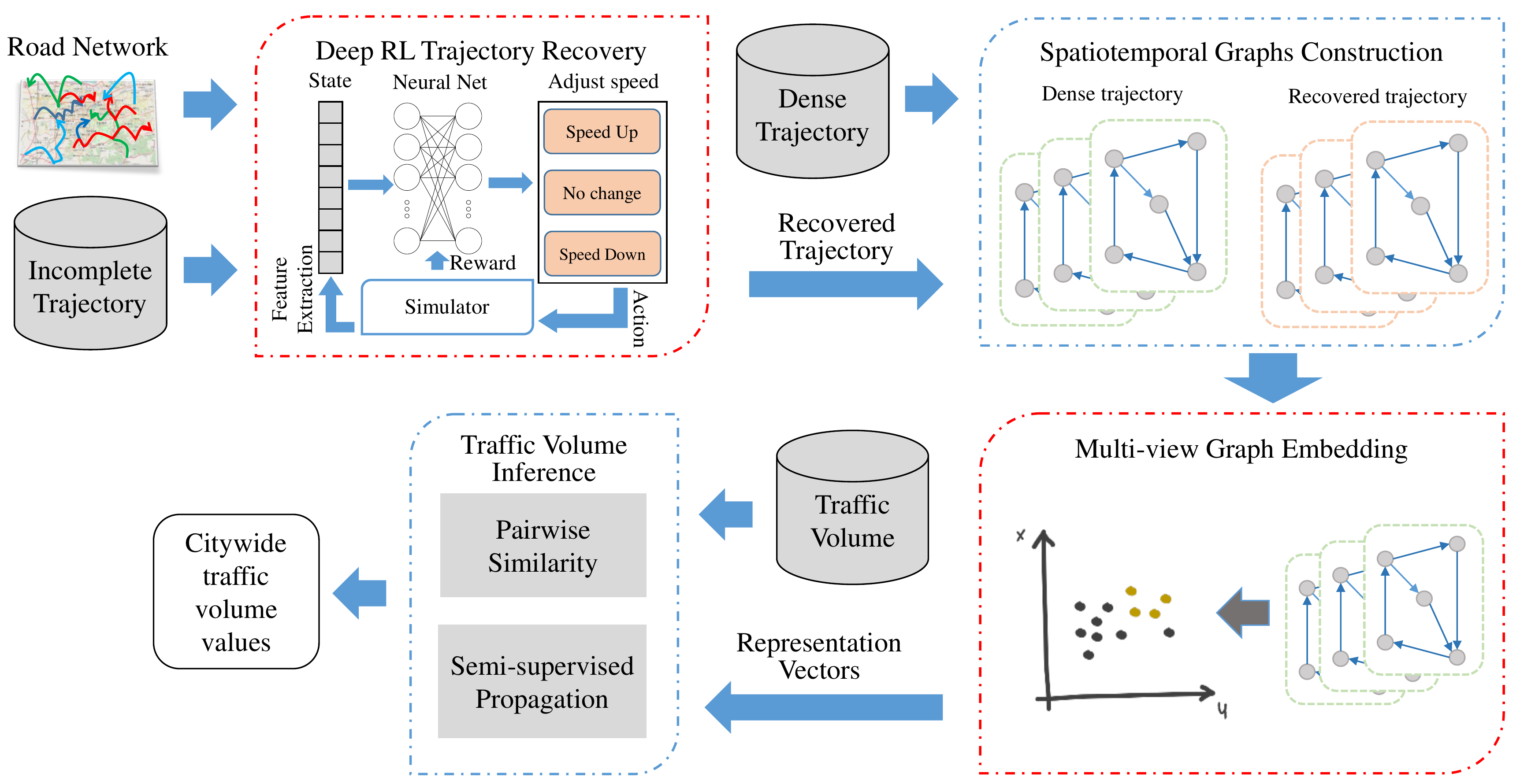}
\caption{Overview of the proposed \ce framework.}
\label{fig.overall}
\end{center}
\end{figure*}

Figure~\ref{fig.overall} illustrates the proposed framework \ce. 
First, we design an algorithm to recover full trajectory from incomplete observations using a traffic simulator, whose parameters are tuned by deep reinforcement learning.
Next, we construct spatiotemporal graphs from the dense and incomplete trajectories. In order to model the spatiotemporal dynamics of traffic volume, we then employ a multi-view graph embedding technique \cite{qu2017attention}  to learn representation vectors for road segments.
Finally, we present a propagation method which infers the citywide traffic volume values according to pairwise similarities between different road segments. In the following sections, we introduce each major component of our solution in detail.

\subsection{Trajectory Recovery}
In this section, we introduce the recovery process for incomplete trajectories. Instead of data-driven algorithms, we incorporate a high-fidelity traffic simulator \cite{sumo2012} for the recovery. Different from previous trajectory recovery methods (e.g., \cite{zheng2012reducing, banerjee2014inferring}), the simulator jointly models the influence of traffic rules as well as vehicle interactions. Furthermore, we propose a novel deep reinforcement learning \cite{mnih2013playing} algorithm that improves the accuracy of trajectory recovery by tuning time-aware hyper-parameters of the simulator.

\subsubsection{Trajectory Recovery via Simulator.}
We first estimate the missing portions of incomplete trajectories by using rich spatiotemporal movement patterns. Unlike the existing methods~\cite{banerjee2014inferring,yang2016efficient,zheng2012reducing,li2015inferring},
our solution relies on a traffic simulator~\cite{sumo2012}, which naturally observes road network constraints (e.g., no turn on red at some intersections) and the global consistency of traffic in the road network. It also captures the influence of vehicle interactions. 

Specifically, we simulate all vehicles simultaneously in real-world road networks using the traffic simulator named \textit{SUMO}~\cite{sumo2012}, conditioning on observations in the incomplete trajectories. Given an incomplete trajectory $\mathbf{\delta}_i = p_{i,1} \rightarrow p_{i,2} \rightarrow \dots \rightarrow p_{i,s}$,  the simulator initializes a vehicle at the corresponding time and location of each point $p_{i,j}$. Then, the vehicle moves toward $p_{i,j+1}$ along the most possible route selected by the simulator. Multiple factors, such as distance, speed, traffic jams, etc., are considered jointly to find the best route. The simulated results for all points $p_{i,j} \rightarrow p_{i,j+1}$ are collected to fill in the missing values in the incomplete trajectories.

However, blindly using the simulator for trajectory recovery without careful validation is not rigorous. Therefore, a novel metric is proposed for trajectory recovery evaluation.
Taking incomplete trajectory $\mathbf{\delta}_i$ as an example, we first measure the relative arrival time $t_{i,j}$ for $p_{i,j} \in \mathbf{\delta}_i$, using time stamps provided by $\mathbf{\delta}_i$. Here $t_{i,j}$ equals to the time consumption of traversing between $p_{i,j}$ and $p_{i,1}$ ($t_{i,1} = 0$) captured by real-world sensors.
Since vehicles would traverse through each monitor point successively in the simulating environment, we compute the relative arrival time at each $p_{i,j}$ again, using time stamps from simulating results. Similarly, let $t_{i,j}^{ \rm simu}$ denote the relative arrival time at $p_{i,j}$ of the simulated vehicle. 
Given those two kinds of relative arrival time $t_{i,j}$ and $t_{i,j}^{ \rm simu}$, the trajectory recovery accuracy can be estimated using the following equation:
\begin{equation}
err = \sum_{\mathbf{\delta}_i \in \mathbf{\Delta}^{\rm I}} \sum_{p_{i,j} \in \mathbf{\delta}_i} \vert t_{i,j}^{ \rm simu} - t_{i,j} \vert / (\sum_{\mathbf{\delta}_i \in \mathbf{\Delta}^{\rm I}} \sum_{p_{i,j} \in \mathbf{\delta}_i} \mathbf{1}).
\label{rlerr}
\end{equation}
The intuition behind Equation \ref{rlerr} is that larger errors will be generated if the recovered trajectories differ a lot from the real ones (i.e., larger $\vert t_{i,j}^{ \rm simu} - t_{i,j} \vert$). And the recovered trajectories have low $err$ if the simulated vehicles arrive at monitor points sharply.

\begin{table}[t]
\caption{Examples of parameters in \textit{SUMO}.}
\centering
\begin{tabular}{|l|l|}
\hline
Parameters & Descriptions \\ \hline\hline
stop & Lets the vehicle stop at the given edge \\ \hline
change lane & Change to the lane with the given index \\ \hline
slow down & Change the speed smoothly\\ \hline
speed & Sets the vehicle speed to the given value\\ \hline
change target & Change the vehicle's destination edge\\ \hline
change route & Assigns the vehicle's new route\\\hline
speed factor & Sets the vehicle's speed factor\\\hline
max speed & Sets the vehicle's maximum speed \\\hline

\end{tabular}
\label{sumopara}
\end{table}

\subsubsection{Deep Reinforcement Learning for Parameter Tuning}
There are a few major parameters  of the simulator (see Table~\ref{sumopara}) we have to tune in order to achieve good simulation results (in other words, small $err$).
For example, changing the speed of vehicles could directly affect the arrival time at each monitored point in the simulating environment, and modifying the lane-changing rules could lead to traffic jams. In order to simulate as realistic as possible, parameter tuning is key.
Unfortunately, it is nontrivial to tune the parameters because they are inter-correlated and their effects on the traffic volume are delayed. For example, we may speed up a vehicle anytime but the results of the acceleration cannot be validated until the vehicle arrives at the next monitor point. Besides, the change of speed of one vehicle may change the speed of others.

To address the above challenges, we propose a reinforcement learning (RL) algorithm to tune the simulator's parameters such that the simulated trajectories can better match the sparsely monitored real data. The RL algorithm successfully avoids the manual configuration of the simulator in a tedious trial-and-error manner.

The general goal of reinforcement learning is to maximize the expected cumulative rewards in a sequential decision making process. Given a set of states $\{s_i\}$ and actions $\{a_i\}$, an action $a$ is taken by the agent following a certain policy $\pi$ after seeing the current state $s$ in each step. The expected return of each potential action $a$ under state $s$ is evaluated using the Bellmen equation \cite{watkins1992q},
\begin{equation}
Q_{\pi}(s,a) = \mathbf{E}(\sum_{i=1}^{\infty}\gamma^{i-1}R_{i} \vert S_0 = s, A_0 = a, \pi),
\label{rlq}
\end{equation}
where $\gamma$ is the discount factor for future rewards. The goal of reinforcement learning is to maximize the expectation of accumulative future rewards through the sequential process.

Parameter tuning of the simulator can be naturally considered as a sequential decision making process, where states are the snapshots of the simulating environment and actions are all possible ways of parameter tuning. We learn a  policy that continuously takes positive action to change the parameters of the simulator in each step. Note that it is intractable to use the Bellmen Equation (Equation \ref{rlq}) to directly optimize the Q-value as it has to track all possible state-action combinations. Instead, we employ Deep Q-Learning (DQN) \cite{mnih2013playing} which approximates the Q-value using a deep neural network.  Given a state $s$, the output of DQN is the approximated Q-values for the actions $Q(s,a;\theta)$, where $\theta$ are neural network weights in DQN. The greedy action is chosen as $\arg\max_{a\in A} Q(s,a;\theta)$. In our context, we use this action to tune the simulator's parameters.

\begin{figure}
\begin{center}
\includegraphics[width=0.46\textwidth]{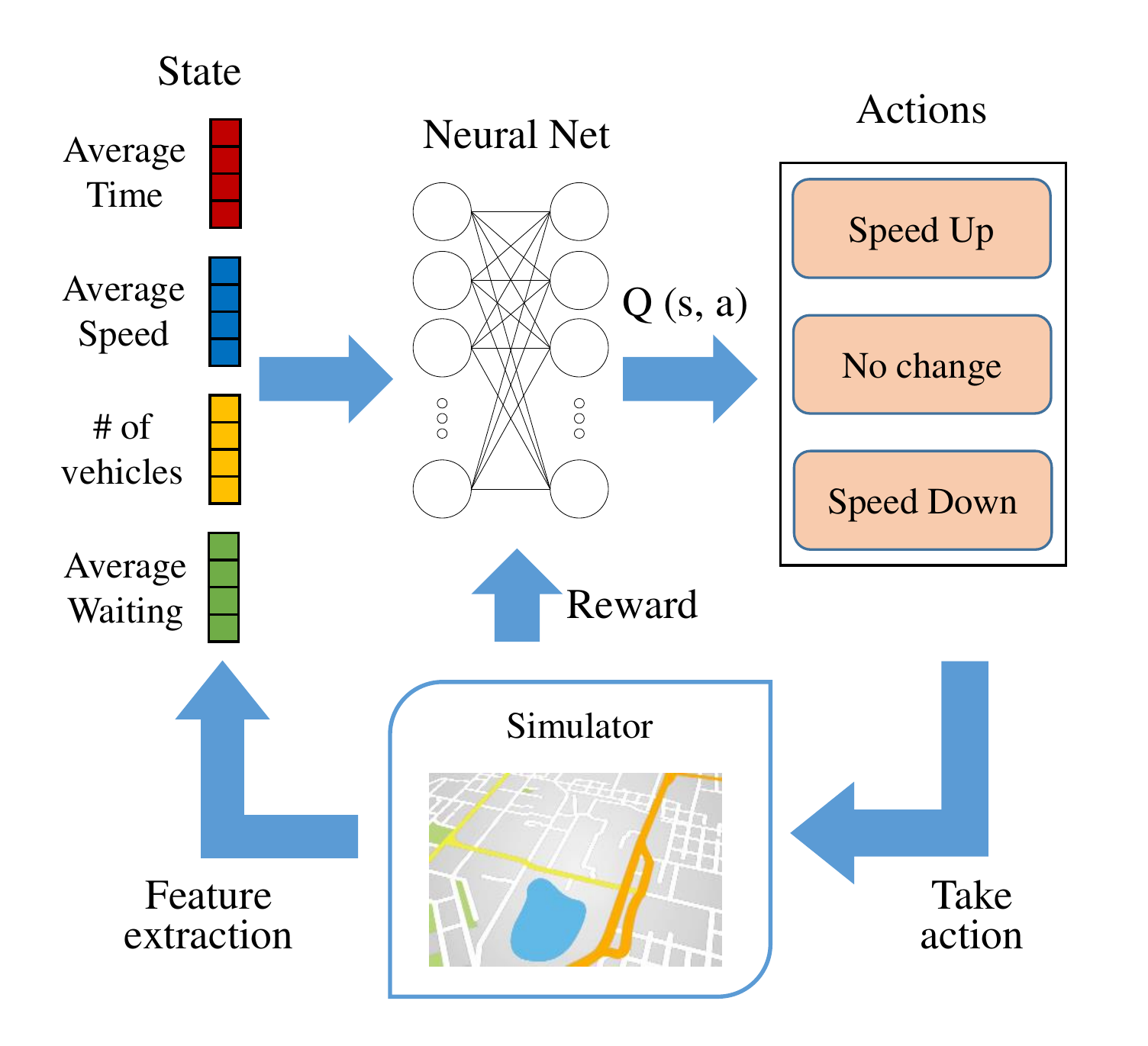}
\caption{Deep reinforcement learning architecture for tuning the vehicles' speeds in the simulator.}
\label{figrl}
\end{center}
\end{figure}

Figure~\ref{figrl} illustrates the DQN framework for trajectory recovery. Real-valued feature vectors extracted from the simulator are used to describe the states. We construct a two-layer fully-connected neural network for DQN to approximate the Q-value. During the training process, weights of the neural network are optimized iteratively to reduce the discrepancy between the optimal Q-value and the predicted one, using gradient descent of the loss function. The details of states, actions, rewards, and the training are described below:

\begin{itemize}
\item \textbf{States}: Many existing deep reinforcement learning methods (e.g. \cite{zheng2018drn, wang2018mathdqn,wei2018intellilight,Li_2018_ECCV}) use real-valued vector to represent state. Following the same practice, we first extract representative features using the API of the simulator. Since the number of vehicles on the road segments directly describes the traffic volume distribution, we first retrieve their values from the simulator. Next, average traversing time and speed are acquired as the traversing speed and time are closely related to the traffic volume. In addition, we retrieve average waiting time, which measures the average stopping time of every vehicle in the immediate past simulating step. Those four types of values are retrieved from each road segment and then concatenated into a real-valued vector as the state representation. Given $m$ road segments, the dimension of the state feature vector is $4m$.

\item \textbf{Actions}: Recall that the action with the largest expected cumulative reward (Q-value) will be selected by the agent in each simulating step. To construct the action set, we change the speeds of vehicles which are one of the most important factors that affect the traversing trajectory. However, directly tuning the speed for every vehicle in the simulator is prohibitive because there could be thousands of vehicles traveling simultaneously. To simplify the setting, we do not change the speed continuously for every vehicle. Instead, the speed limits (meters/second) for different vehicles  are modified within a certain range in each simulation step. The actions include increasing the speed limit by $+1$, $0$, or $-1$ (i.e., decrease by $1$) for each vehicle. To reduce the action space, we group the vehicles according to their sizes (i.e., sedans, SUVs, and large trucks) and assign the same speed limit to a group of vehicles. Hence, the action space consists of $9$ possible actions pairing the speed limits with the groups. Such a setting actually changes the behavior of every group of vehicles by limiting their speeds into a certain range.

\item \textbf{Rewards}:
In reinforcement learning, positive or negative rewards are provided by the environment for each action by the agent. The agent can utilize the reward to calculate the optimal Q-value, and to optimize parameters in the neural network using the discrepancy induced by the Bellman equation.
We construct short-term rewards according to the following equation:
\begin{equation}
R = \sum_{v \in \Omega}(e^{-\vert t_v^{\rm simu} - t_v\vert}),
\label{rlr}
\end{equation}
where $\Omega$ denotes vehicles that have arrived at the monitor points at the simulating step. That is, the simulating results of those vehicles can be evaluated at this time. The arrival time stamps $t_v^{simu}$ and $t_v$ are of the simulated ones and the real vehicles, respectively. 
The definition is straightforward. Obviously, lower absolute time differences are received if and only if more vehicles arrive sharply, and the agent receives larger rewards under such conditions.

\item \textbf{Training}: A simulating step is set to lasting for one minute. A two-layer neural network is initialized and trained to estimate the Q-value. Weights $\theta$ of the network are updated every step by gradient descent, using the discrepancy between estimated Q-value and the feedback from the simulating environment. Before the training, we first set up a replay memory $\mathcal{M}$ to store state transitions (i.e., $(s, a, s^\prime, R)$ where $s^\prime$ denotes the next state following $s$ after taking action $a$). The deep reinforcement learning algorithm keeps sampling mini-batches from the replay memory and updates its network weights in each simulating step by minimizing the following loss function:
\begin{equation}
\mathcal{L}(\theta) = \mathbf{E}_{s,a}[(Q^{*} - Q(s,a;\theta))^2],
\label{rlloss}
\end{equation}
where $Q^{*} = R + \gamma \max_{a^\prime} Q(s^\prime, a^\prime; \theta)$ is the target optimal Q-value, which is the sum of the short-term reward $R$ and the optimal Q-value of the next step. The decay term $\gamma$ gives penalty to future rewards. The expectation is computed over the mini-batches. Furthermore, we adopt $\epsilon$-greedy strategy \cite{mnih2015human} to balance the trade-off between exploration and exploitation. That is, with probability $\epsilon$, a random action is selected for exploration during training, instead of the optimal one.
After calculating the loss function, the network weights $\theta$ are updated using gradient descent. The gradient of $\theta$ is given as follows,
\begin{equation}
\nabla_{\theta} \mathcal{L}(\theta) = \mathbb{E}_{s,a}[(Q^{*} - Q(s,a;\theta)) \nabla_{\theta} Q(s,a;\theta)].
\label{rlgrad}
\end{equation}

\end{itemize}

The overall training procedure of the  deep reinforcement learning architecture is given in Algorithm \ref{rlalg}.
Once the simulator is stable, we fill in the missing values of the real and incomplete trajectories using the traffic volume values in the simulator \textit{SUMO}. Compared with taxi trajectories collected from GPS devices, these recovered trajectories, denoted by $\Delta^{\rm I+}$, are more representative of a variety of vehicle types. Both kinds of trajectories will be employed in the following steps.

\begin{algorithm}[t]
	\caption{Training of the Deep RL Architecture}
    \label{rlalg}
	Randomly initialize network parameters $\theta$ and the replay memory $\mathcal{M}$\;
	\For{each episode}{
	    \For{each simulation step}{
	        \tcc{Action and simulating}
	        \eIf{random() < $\epsilon$}
	        {
	            Select a random action $a$\;
	        }
	        {
	            Select $a = \max_{a} Q(s, a; \theta)$\;
	        }
	        Obtain short-term reward $R$ and new  state $s^\prime$ by executing $a$ in the simulator\;
	        Store state transition $(s, a, s^\prime, R)$ in $\mathcal{M}$\;

	        \tcc{Updating Q-network parameters $\theta$}
	        Sample mini-batch state transitions $\{(s, a, s^\prime, R)\}$ from  $\mathcal{M}$\;
	        \eIf{simulator terminated}
	        {
	            $Q^{*} = R$\;
	        }{
	            $Q^{*} = R + \gamma \max_{a^\prime} Q(s^\prime, a^\prime; \theta)$\;
	        }
	        Use Equation \ref{rlgrad} to perform gradient descent on the loss $\mathcal{L}(\theta)$ based on $Q^*$\;
	    }
	}
\end{algorithm}

\subsection{Spatiotemporal Graph Construction}
\begin{figure}
\begin{center}
\includegraphics[width=0.43\textwidth]{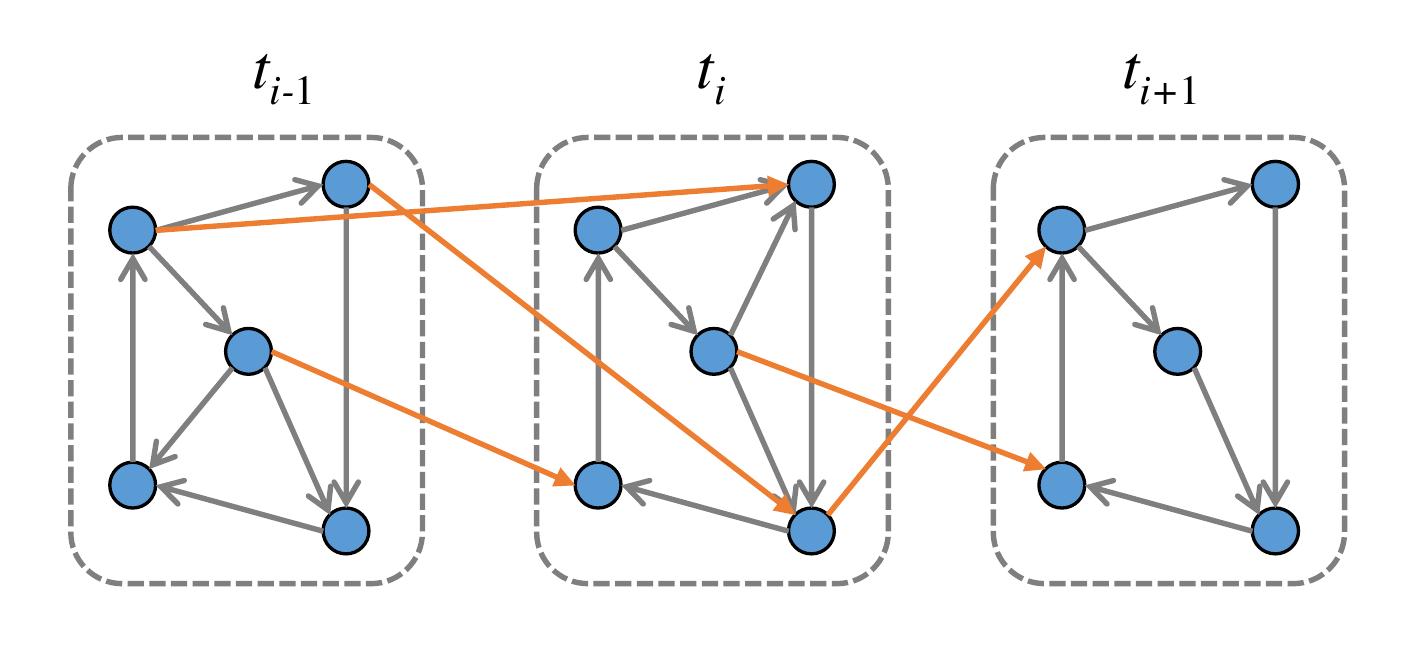}
\caption{An example of spatiotemporal graph. Three layers are shown in the figure. Gray arrows denote edges within the same layer and yellow arrows denote edges between current and next layers.}
\label{stgraph}
\end{center}
\end{figure}

Temporal correlations of different road segments change over time. For example, an expressway connecting business areas and residential areas may exhibit reversed traffic flows from morning to evening. Moreover, the spatial relationship between road segments is extremely complicated (e.g., overpasses at the intersection of the main road). To model such dynamism of traffic in the road network, we construct spatiotemporal graphs, as shown in Figure \ref{stgraph} and detailed below.

A spatiotemporal graph in this paper is a multi-layer graph $G = \langle\mathcal{V},\mathcal{E}\rangle$, where $\mathcal{V} = \{v_i^t\}$ denotes time-enhanced nodes representing road segments $\{r_i\}$, and $\mathcal{E}$ represents edges between nodes. Nodes in the same time interval are grouped to the same \textit{layer}. To preserve spatial restrictions of the road network, edges in $\mathcal{E}$ are limited to $v_i^t \rightarrow v_j^{\{t+1\}} $ where vehicles could travel from the road segment $r_i$ to $r_j$.  The edge's weight is assigned using the traffic flow value, i.e., number of trajectories traversing through the edge. 

To differentiate mobility behaviors and traffic patterns in different sources of trajectories, two spatiotemporal graphs are constructed respectively using the dense trajectories $\Delta^{\rm D}$ and recovered $\Delta^{\rm I+}$ from incomplete trajectories.

\subsection{Multi-view Graph Embedding}
The objective of citywide traffic volume inference is to estimate the unknown traffic volume values from the observed ones. This problem is challenging due to multiple interwoven factors such as the road network, vehicles' physical condition, road condition, time, etc. Our solution relies on the spatiotemporal graphs constructed from the macro trajectories, effectively factoring out those individual factors and allowing us to conduct the inference on the graphs instead. 

We first embed the time-enhanced road segments (nodes of the graph) to real-valued vectors using the two spatiotemporal graphs obtained above.
The embedding captures multi-hop correlations~\cite{Wang2017} between road segments at different times. Moreover, joint embeddings from the two graphs naturally account for both the GPS-tracked taxi and other vehicles monitored in the incomplete trajectories. We first introduce the embedding process on a single graph, followed by the joint embedding algorithm from the two graphs by a multi-view graph embedding method~\cite{qu2017attention}.

We adopt the Skip-gram model \cite{mikolov2013efficient} to learn the embedding vector $\mathbf{u}_i^t$ for each node $v_i^t$ in a spatiotemporal graph $G = \langle\mathcal{V},\mathcal{E}\rangle$. 
Given a node in the graph, the skip-gram model aims to predict the probability of the node's context.
We first define a path in $G$ as $P_i = v_{i_1}v_{i_2}\dots v_{i_\sigma}$, where $v_{i_j}$ in $P_i$ is corresponding to a node in $G$. We omit the time superscript here, which can be easily acquired from each layer. Given a specific node $v_i^t$, the set of all paths containing $v_i^t$ formulates a set $\mathcal{P}(v_i^t)$.
We define set $\mathcal{N}(v_i^t) = \{v_c\vert \exists P_i \in \mathcal{P}(v_i^t), v_c \in P_i\} \backslash \{v_i^t\}$ as the context of node $v_i^t$, which refers to all multi-hop neighbors of $v_i^t$ (i.e., nodes closely surround $v_i^t$ in space and time) \cite{wang2019simple}.
Specifically, the following log-probability is maximized:
\begin{equation*}
\begin{aligned}
\mathcal{O} & = \sum_{v_i^t \in G} \log Pr(\mathcal{N}(v_i^t) \vert v_i^t))
= \sum_{v_i^t \in G} \log \prod_{v_c \in \mathcal{N}(v_i^t)} P(v_c \vert v_i^t)
\\ &= \sum_{v_i^t \in G}  \sum_{v_c \in \mathcal{N}(v_i^t)} \log P(v_c \vert v_i^t).
\end{aligned}
\end{equation*}
and $P(v_c \vert v_i^t)$ is further estimated by 
$
P(v_c \vert v_i^t) = \sigma({\mathbf{u}_i^t}^{\rm T} \mathbf{u}_c^\prime)
$
where $\sigma(\cdot)$ is the sigmoid function, and $ \mathbf{u}_c^\prime$ denotes the auxiliary vector of $v_c$ when it is treated as ``context'' of other nodes.

To achieve better performance and reduce computational costs, negative sampling technique \cite{mikolov2013distributed} is applied. For each node $v_i^t$ and a neighbor node $v_c$, we sample negative nodes $NG(v_i^t) = \{z\vert z\notin \mathcal{N}(v_i^t)\}$ which denotes unrelated road segments.
The objective function for network embedding with negative sampling is:
\begin{equation*}
\begin{aligned}
& \mathcal{O}  =  \sum_{v_i^t \in G}  \sum_{v_c \in \mathcal{N}(v_i^t)} \Big\lbrace \log P(v_c \vert v_i^t) + \sum_{z \in NG(v_i^t)}\log(1 -  P(v_c \vert v_i^t)) \Big\rbrace
\\ & =  \sum_{v_i^t \in G}  \sum_{v_c \in \mathcal{N}(v_i^t)} \Big\lbrace \log (\sigma({\mathbf{u}_i^t}^{\rm T} \mathbf{u}_c^\prime)) + \sum_{z \in NG(v_i^t)}\log(1 -  \sigma({\mathbf{u}_i^t}^{\rm T} \mathbf{u}_z^\prime)) \Big\rbrace.
\end{aligned}
\end{equation*}
The objective reads that, for each node $v_i^t$, we aim to maximize the probability of its co-occurrence with nodes in $\mathcal{N}(v_i^t)$, while minimize the probability of it being neighbor with nodes from $ NG(v_i^t)$.

Since we have two spatiotemporal graphs of the same set of road segments, we may learn the embedding by a trade-off between the objective functions above over the two graphs. Namely,
\begin{equation}
\mathcal{O} = \alpha \mathcal{O}_{G^{\rm D}} + (1-\alpha) \mathcal{O}_{G^{\rm I+}},
\end{equation}
where $G^{\rm D}$ and $G^{\rm I+}$ are the graphs constructed from dense  and recovered trajectories, respectively, and $\alpha$ balances the two terms.

\subsection{Citywide Traffic Volume Inference}

After learning representations for road segments that preserve spatiotemporal dynamics, we borrow a semi-supervised learning formulation~\cite{zhu2003semi} to propagate the traffic volume values to the unknown road segments.
At first glance, one may train a regression model taking as input the road segment's embedding to predict the missing traffic volume data. However, such methods fail when  only a small portion of the entire road network is monitored. We instead employ a formulation which is popular in semi-supervised learning for the inference.

We first define pairwise similarities between any two road segments. The similarity of $r_i$ at time interval $t$ and $r_j$ at time interval $t^\prime$ is measured by:
\begin{equation}
sim(r_{i}^{t}, r_{j}^{t^\prime}) := \omega_{i,j}^{t, t^\prime} = \langle \mathbf{u}_{i}^{t} \cdot \mathbf{u}_{j}^{t^\prime} \rangle,
\label{oldsimi}
\end{equation}
where $\langle \cdot \rangle$ denotes the inner product of two vectors.

We then prune out irrelevant pairs of road segments by constructing 0-1 masks based on geospatial features (i.e., absolute distance between road segments) as well as temporal distance. The intuition behind such pruning is that traffic volume values of faraway locations usually do not have similarities, and the temporal dependencies of traffic pattern at the same location will decrease as the time lag becomes longer \cite{yao2018deep}. The ``masked'' pairwise similarities between road segments are defined as:
\begin{equation}
{\omega^{\star}}_{i,j}^{t, t^\prime} = \left\{
  \begin{array}{@{}ll@{}}
    \omega_{i,j}^{t, t^\prime}, & \text{if}\ adj(r_i,r_j) \land \vert t - t^\prime \vert \leq 1, \\
    0, & \text{otherwise.}
  \end{array}\right.
\label{newsimi}
\end{equation}
Here $adj(r_i,r_j)$ is $true$ if and only if   $r_i$ and $r_j$ are adjacent on the road network. The above mask ensures only the correlations between road segments that are close enough in time and space are considered. The citywide traffic volume values $\{x_{it}\}$ are inferred by minimize the following with respect to (w.r.t.) the unknown values:
\begin{equation}
\mathcal{L}^{\star} = \sum_{\substack{1 \leq i, j \leq m, 1 \leq t, t^\prime \leq n}} {\omega^{\star}}_{i,j}^{t, t^\prime} (x_{it} - x_{jt^\prime})^2.
\label{obj}
\end{equation}
Since the objective function is convex w.r.t.\ the unknown traffic volume values, gradient descent is applied to find the global solutions.

\section{Experiments}

In this section, we conduct extensive experiments to answer the following research questions.

\begin{itemize}
    \item How does the proposed \ce perform compared with the baseline methods?
    \item How do each component of  \ce affect the overall performance?
    \item How accurate is the trajectory recovery via the deep RL architecture?
    \item How to balance different types of trajectories through the hyper-parameter $\alpha$?
    \item How is the inference error distributed over space (i.e., different road segments) and time (i.e., different time periods of a day)?
\end{itemize}

\subsection{Datasets}
We compile a traffic volume dataset using the vehicle counting reported by the city surveillance system from a large provincial capital city in China. The incomplete trajectories are acquired using plate numbers identified by surveillance cameras (with an average F1-score of 90\%). Dense trajectories are collected from GPS sensors equipped by public taxicabs. The average reporting frequency of the taxicab-carried GPS devices is 5 seconds. All those datasets are collected over the period of Aug.\ 2016. 
The road network is extracted from \textit{Openstreetmap.org}  \cite{OpenStreetMap}. 
We split roads into segments using intersections as natural cut-points and set the length of the time interval to 5 minutes. Moreover, geospatial features such as starting/ending locations, road segment length, road width, number of lanes, road type, speed limit, etc., are constructed for each road segment. We select two regions in the city. The detailed data descriptions are shown in Table~\ref{setting}.

\subsection{Experiment Settings}
We cut trajectories recursively at any $p_{i,j}\rightarrow p_{i,j+1}$ if the time interval between $p_{i,j}$ and $p_{i,j+1}$ is larger than $30$ minutes. Because vehicles may stop and park during such long duration, their behaviors are different from the constantly moving vehicles.

We randomly split the road segments with traffic information into training (80\%) and testing (20\%), respectively. We further select 20\% of the training randomly as validation. The same split is used for all experiments. We run each baseline and ablation 10 times to report the average results.

We set the memory buffer size to $10,000$ and the discount factor $\gamma$ to $0.8$ for the deep RL framework. A feed-forward neural network with two $256$-dimension hidden layers is trained using the Adam optimizer \cite{kingma2014adam}. The size of a mini-batch is $128$ and the learning rate is $0.0001$. Parameters of the network are updated at every step of the simulator. 
The probability $\epsilon$ in the deep RL training is reduced linearly from $0.5$ to $0.01$ over $2,000$ epochs. The speed limits for every type of vehicle are restricted between $1m/s$ and $40m/s$ in the simulating environment.
We set the dimension size for embedding vectors to $50$. The window size for sampling in Skip-gram is $10$.

\begin{table}
\caption{Dataset statistics.}
\centering
\begin{tabular}{|l|l|c|c|}
\hline
\multicolumn{2}{|l|}{ } & Region 1 & Region 2 \\ \hline\hline
\multicolumn{2}{|l|}{Time period} & Aug. 1 - Aug 4 & Aug. 14 - Aug. 20 \\ \hline
\multicolumn{2}{|l|}{Time interval} & 5 minutes & 5 minute \\ \hline
\multicolumn{2}{|l|}{Longitude range} & 116.8988 - 116.9457 & 117.0635 - 117.1053 \\ \hline
\multicolumn{2}{|l|}{Latitude range} & 36.6507 - 36.6926 & 36.6624 - 36.7052 \\ \hline
\multicolumn{2}{|l|}{\# of  all road seg.} & 322 & 143 \\ \hline
\multicolumn{2}{|l|}{\# of monitored seg.} & 45 & 32 \\ \hline
\multicolumn{2}{|l|}{\# of incomplete traj.} & 366,645 & 1,798,782 \\ \hline
\multicolumn{2}{|l|}{\# of dense taxi traj.} & 12,562 & 97,315 \\ \hline
\end{tabular}
\label{setting}
\end{table}

\subsection{Metrics}
We use Root Mean Square Error (RMSE) and Mean Absolute Percentage Error (MAPE)  to evaluate the performance of all methods.
The detailed definitions of the two metrics are stated as below:

$$RMSE=\sqrt{\frac{1}{s}\sum_{l=1}^{s}(x_l-\hat{x}_{l})^2},\ \ MAPE=\frac{1}{s}\sum_{l=1}^{s}\frac{|x_l-\hat{x}_{l}|}{\hat{x}_{l}},$$
where $x_l \in \{x_{it}\}$ is a test sample, $\hat{x}_{l}$ is the ground truth of $x_l$, and $s$ denotes the total number of test samples.

While RMSE receives more penalties from larger values, MAPE focuses on the error of smaller values. Therefore, combining these two metrics can lead to more comprehensive conclusions.
Similar to~\cite{yao2018deep,yao2019revisiting}, to evaluate relative error using MAPE, we filter out samples whose  traffic volume  is lower than $5$ in testing, since road segments with very low traffic volume are of little interest for our problem.

\subsection{Baselines}
Our solution is compared with the following baselines: (1) two geospatial-based methods, (2) two supervised learning methods, and (3) two semi-supervised learning methods.

\begin{itemize}
\item \textbf{Spatial \textit{k}NN} selects top $k$ nearest  road segments with traffic data and uses the average of their volume values at each time interval as the traffic volume prediction for unmonitored road segments.

\item \textbf{Contextual Average} (CA)  uses the averaged volume values from same-type road segments (e.g., primary, expressway, etc.) as the prediction. 

\item \textbf{Linear Regression} (LR) is trained on all road segments with traffic volume using geospatial features. We train regression models for each time interval.

\item \textbf{XGBoost} \cite{chen2016xgboost} is a boosting-tree-based method which is popular in data mining community. We train XGBoost separately on each time interval.

\item \textbf{Graph SSL} Classic graph-based semi-supervised learning method with loss function $\mathcal{L} =  \sum_t  \sum_{i,j} a_{i,j} (x_{it}-x_{jt})^2 $, which is implemented following \citeauthor{zhu2003semi}. The similarity weight $a_{i,j}$ is defined using the Euclidean distance between the geospatial feature vectors $r_i$ and $r_j$.
 
\item \textbf{ST-SSL} \cite{meng2017city} is the state-of-the-art method for inferring citywide traffic volume using dense trajectories. ST-SSL utilizes static features and estimated speed patterns from taxi trajectories to model the correlations between road segments. The citywide traffic volume is then inferred by a semi-supervised learning method.

\end{itemize}

\subsection{Result Analysis of Different Methods}
The comparison of \ce with other methods in two regions is shown in Table \ref{result}. As we can see, \ce achieves best performances in both regions.
In particular, \ce significantly outperforms geospatial-based methods due to its capability of well modeling the dynamism of traffic volume.
In addition, \ce achieves tremendous improvement compared with supervised methods. This is because supervised methods are trained on road segments with traffic data, they ignore correlations between other road segments. 
Moreover, both geospatial and regression-based methods fail to model spatiotemporal patterns in multi-sources trajectories, and do not well utilize information from unmonitored road segments.

\ce significantly outperforms both Graph SSL and ST-SSL. There are two potential reasons. First, \ce  utilizes vast incomplete trajectory data to recover spatiotemporal similarities among road segments. The proposed deep reinforcement learning recovery method successfully transforms partial observations to complete route samples. Second, the joint embedding algorithm enhances the representation of road segments. Both multi-hop correlations and multi-view graph information are preserved in learned embedding vector, resulting in better performance. Furthermore, Graph SSL achieves lower errors compared with ST-SSL, a possible explanation is that the strong assumption of speed similarity does not hold in those regions. For example, most drivers will drive close to the speed limit unless the road capacity reaches the upper limit. In such condition, speeds are almost stationary while traffic volume could change depending on the number of vehicles. Moreover, it suffers from the spatial sparsity of taxi trajectories since the coverage of taxi trajectory in the city is limited.

\begin{table}[t]
\caption{Comparison of baseline methods.}
\centering
\begin{tabular}{|l|c|c|c|c|}
\hline
\multicolumn{1}{|c|}{\multirow{2}{*}{Methods}} & \multicolumn{2}{c|}{Region 1} & \multicolumn{2}{c|}{Region 2} \\ \cline{2-5} 
\multicolumn{1}{|c|}{} & RMSE & MAPE & RMSE & MAPE \\ \hline \hline
Spatial kNN & 7.1404 & 0.6348 & 8.5259 & 0.6053 \\ \hline
CA & 8.0866 & 0.5251 & 8.1317 & 0.5423 \\ \hline
LR & 13.2857 & 1.0911 & 11.5781 & 1.0219 \\ \hline
XGBoost & 9.2157 & 0.6711 & 9.2847 & 0.6954 \\ \hline
Graph SSL & 6.1743 & 0.4457 & 6.3640 & 0.4643 \\ \hline
ST-SSL & 6.5603 & 0.5631 & 7.1528 & 0.5975 \\ \hline
\ce (Ours) & \textbf{5.5877} & \textbf{0.3675} & \textbf{5.6680} & \textbf{0.3915} \\ \hline
\end{tabular}
\label{result}
\end{table}

\subsection{Ablation Study}
To validate each component of \ce, the following variations of \ce are further studied. Despite the changed part(s), all variations have the same framework structure and parameter settings. 
The results of all variations are listed in Table \ref{varresult}.

\begin{itemize}
\item  \textbf{\ce-Reg} trains XGBoost models using representation of road segments. It can be considered as appending embedding vectors to the geospatial features in previous XGBoost.

\item  \textbf{\ce-Tr} replaces the trajectory recovery part using algorithm in \cite{banerjee2014inferring}. The method infers uncertain trajectories from network-constrained partial observations by summarizing all probable routes in a holistic manner.

\item \textbf{\ce-Df} recovers trajectories using default simulator, i.e., we directly collect results from \textit{SUMO} without training the deep reinforcement learning component.

\item \textbf{\ce-Semi} directly applies graph-based semi-supervised learning \cite{zhu2003semi} on the spatiotemporal graphs to propagate known traffic volume without the embedding process. The similarity between any time-enhanced road segments (nodes of the graph) is assigned according to the edge weights (i.e., traffic flow value from trajectory data).

\item \textbf{\ce-Um} proceeds semi-supervised learning without pruning out irrelevant pairs of road segments. Namely, the raw similarity in Equation (\ref{oldsimi}) is used.

\end{itemize}

\ce-Reg performs the worst among all variations. However, it still outperforms the original XGBoost. In addition, the performance gap between \ce-Reg and \ce reflects the advantage of the semi-supervised method, which successfully leverages large amounts of unmonitored road segments.

The comparison between \ce-Tr and \ce highlights the effectiveness of the proposed trajectory recovery method. Sampling-based methods are unable to handle complex road network constraints, vehicle interactions, and other factors, thus fail to recovery highly accurate trajectories.
Moreover, \ce-Df performs worse compared with \ce. A possible explanation is that default stationery (time independent) parameters do not describe real-world vehicle movements correctly. Furthermore, since there are fewer vehicles in the simulator, some parameters should still be altered even if the ground truth is known. For example, vehicles traverse faster and arrive earlier in a simulated environment, resulting in a higher error of trajectory recovery. In such condition, speed limits could be explicitly reduced to adjust vehicles' movement.

The performance of \ce-Semi is obviously overtaken by \ce. This is because the joint embedding enhances the representation capacity from multi-hop spatiotemporal correlations and multi-view graph information.

The comparison of \ce-Um and \ce indicates the effectiveness of proposed 0-1 masks that explicitly remove unrelated road segment pairs. The propagation of traffic volume focuses on relevant pairs of road segments, improving the overall performance of \ce distinctly.

\begin{table}[t]
\caption{Comparison of variations.}
\centering
\begin{tabular}{|l|c|c|c|c|}
\hline
\multicolumn{1}{|c|}{\multirow{2}{*}{Methods}} & \multicolumn{2}{c|}{Region 1} & \multicolumn{2}{c|}{Region 2} \\ \cline{2-5} 
\multicolumn{1}{|c|}{} & RMSE & MAPE & RMSE & MAPE \\ \hline \hline
{\ce-Reg} & 7.0851 & 0.5182 & 6.9187 & 0.5155 \\ \hline
{\ce-Tr} & 6.5091 & 0.5127 & 6.8262 & 0.5390 \\ \hline
{\ce-Df} & 6.1445 & 0.4104 & 6.4006 & 0.4378 \\ \hline
{\ce-Semi} & 7.9121 & 0.5673 & 7.5165 & 0.5366 \\ \hline
{\ce-Um} & 6.0357 & 0.3781 & 6.2272 & 0.4057 \\ \hline
{\ce}   & \textbf{5.5877} & \textbf{0.3675} & \textbf{5.6680} & \textbf{0.3915} \\ \hline
\end{tabular}
\label{varresult}
\end{table}

\subsection{Accuracy of Trajectory Recovery}
We evaluate the trajectory recovery algorithm. Since the spatial ground truth for incomplete trajectories is unavailable, we construct validation from arrival times at monitor points.
First, we manually tune speed limit parameters using greedy search approach over each hour of the simulating procedure to minimize the arrival time error. Those speed parameters are used as the default setting of the compared method. Note that we do not use the default speed limit parameters in \textit{SUMO} (i.e., $22.4m/s$).
Following Equation~(\ref{rlerr}), the averaged arrival time  error (in second) of the default simulator and that of the optimized simulator are compared.
The results are demonstrated in Table \ref{rlresult}. The proposed deep reinforcement learning method reduces averaged arrival error significantly, which is relatively lower by 18.29\% in Region 1 and 13.61\% in Region 2. Moreover, the improvement in Region 1 is relatively higher than in Region 2. A possible reason is that Region 1 has a relatively simpler environment with less traffic. Therefore, modeling the state-action relationship could be easier for the DQN.

\begin{table}[t]
\caption{Averaged arrival time error ($s$) of simulated vehicles.}
\centering
\begin{tabular}{|l|c|c|}
\hline
Region & Region 1 & Region 2 \\ \hline\hline
Default Parameters & 27.2356 & 29.2141 \\ \hline
Deep RL & \textbf{22.2554} & \textbf{25.2231} \\ \hline
\end{tabular}
\label{rlresult}
\end{table}

\subsection{Hyper-Parameter}
\begin{figure}[!t]
\centering
\begin{subfigure}[b]{.23\textwidth}
  \centering
  \includegraphics[width=\columnwidth]{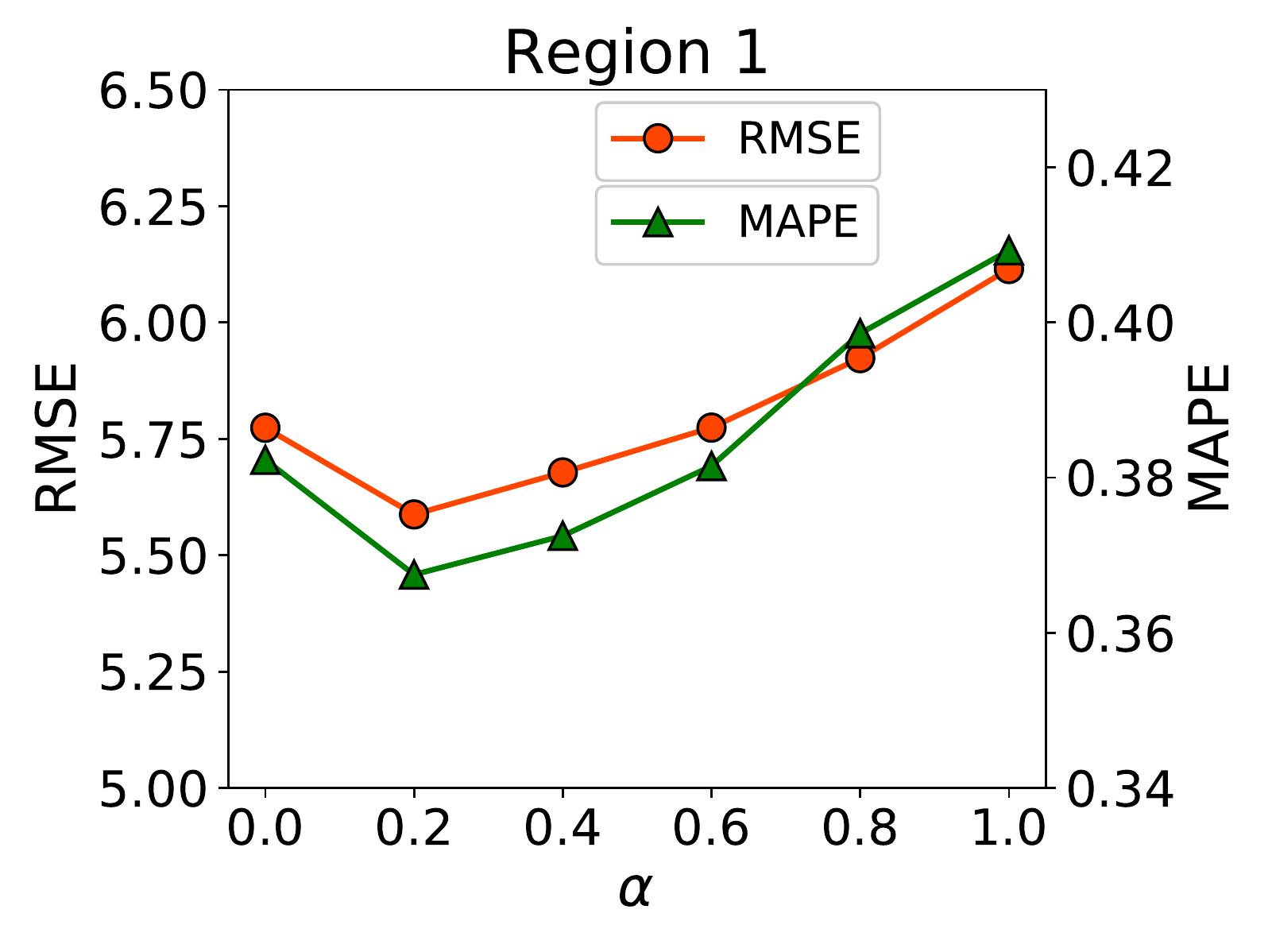}
\end{subfigure}
\begin{subfigure}[b]{.23\textwidth}
  \centering
  \includegraphics[width=\columnwidth]{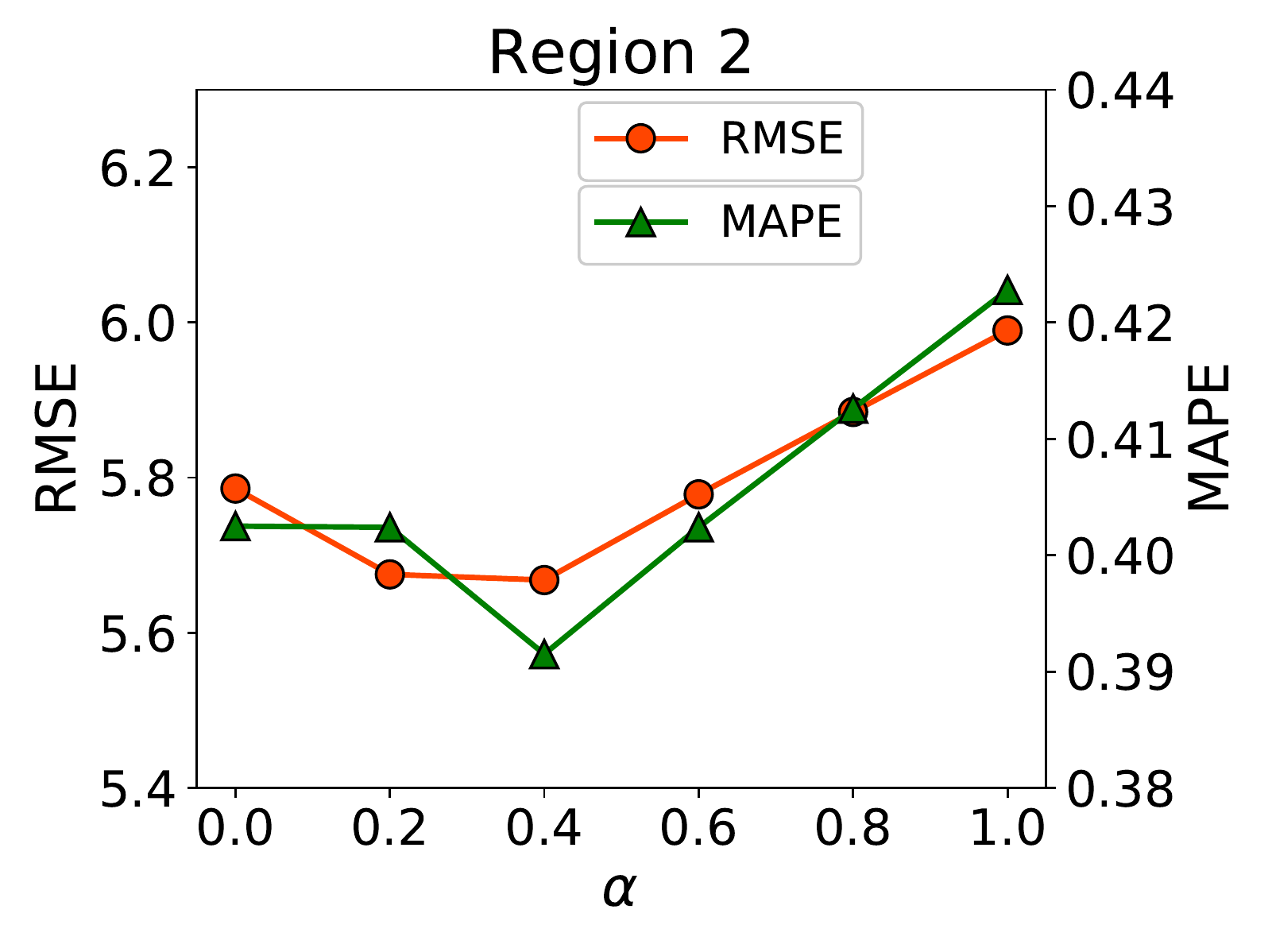}
\end{subfigure}
\caption{Performance w.r.t different $\alpha$.}
\label{alphastudy}
\end{figure}

We further study the hyperparameter $\alpha$ which balances weights between different spatiotemporal graphs.  More emphasis is put to  recovered trajectories when $\alpha$ is closer to $0$, and vise versa when $\alpha$ is closer to $1$. The performance of \ce w.r.t different $\alpha$ is shown in Figure \ref{alphastudy}. As we can observe, the inference error first drops and then rises in both regions. Moreover, \ce achieves relatively low inference errors when $\alpha$ falls into a certain range. However, the performance is worse when $\alpha$ is closer to $0$ or $1$.
In addition, the performance of \ce is significantly improved in both regions by leveraging incomplete trajectories. Compared with the ablation that only uses taxi data (i.e., $\alpha=1$), \ce reduces the RMSE value by 8.84\%, 5.47\%, and the MAPE value by 9.76\%, 7.41\%  in region 1 and region 2, respectively. This suggests the importance of modeling multi-source trajectories jointly. Besides, the relative improvement from recovered trajectories in Region 1 is larger compared with Region 2. A potential explanation is that the taxi trajectories in Region 1 are very sparse, thus brings biased spatiotemporal correlation and results in larger inference error. Moreover, recovered trajectories also contain noises and uncertainties, and that's why  \ce achieves the best performance by combining them together.

\subsection{Error Distribution Interpretation}
We also explore the spatiotemporal performance of \ce, i.e., error distribution over different types of roads, and over different time periods of a day. 
Without loss of generality, Region 2 is selected for the interpretation.

\subsubsection{Error distribution over the space.}
\label{spaterr}
To study the performance of \ce over different road segments, we first split road segments from testing set into two categories using road type information extracted from the map. Namely, Group 1 (i.e., major roads such as highway and primary way) and Group 2 (secondary roads like tertiary route). Generally, road segments from Group 1  have more lanes and higher average traffic volume values. 

Table \ref{sttable} provides a detailed illustration of the spatial performance of \ce. As we can see, the RMSE values are higher for road segments from Group 1, as they have higher traffic volume values. RMSE values are positively correlated to the original variables themselves. Meanwhile, we also observe that the MAPE values of road segments from Group 2 is relatively lower than those of road segments from Group 1. One potential reason is that traffic volume of road segments from Group 1 has higher uncertainty. For example, traffic jams occur when the traffic volume exceeds road capacity, and this usually happens on major roads. Moreover, local events have a higher probability to happen near major roads. Those factors increase the uncertainty of traffic volume on road segments from Group 1, resulting in a high relative error (i.e., MAPE) of the framework.

\begin{table}[]
    \caption{Error distribution over different groups of road segments.}
    \centering
    \begin{tabular}{|l|c|c|c|c|}
    \hline
    \multicolumn{1}{|c|}{\multirow{2}{*}{Group}} & \multicolumn{2}{c|}{Region 1} & \multicolumn{2}{c|}{Region 2} \\ \cline{2-5} 
    \multicolumn{1}{|c|}{} & RMSE & MAPE & RMSE & MAPE \\ \hline \hline
    Group 1 & 6.5151 & 0.4102 & 6.6741 & 0.4170 \\ \hline
    Group 2 & 3.5488 & 0.3368 & 3.4101 & 0.3274 \\ \hline
    \end{tabular}
    \label{sttable}
\end{table}

\subsubsection{Error distribution over different time periods of the day.}
\begin{figure}
    \begin{subfigure}[b]{.45\textwidth}  
        \centering 
        \includegraphics[width=\columnwidth]{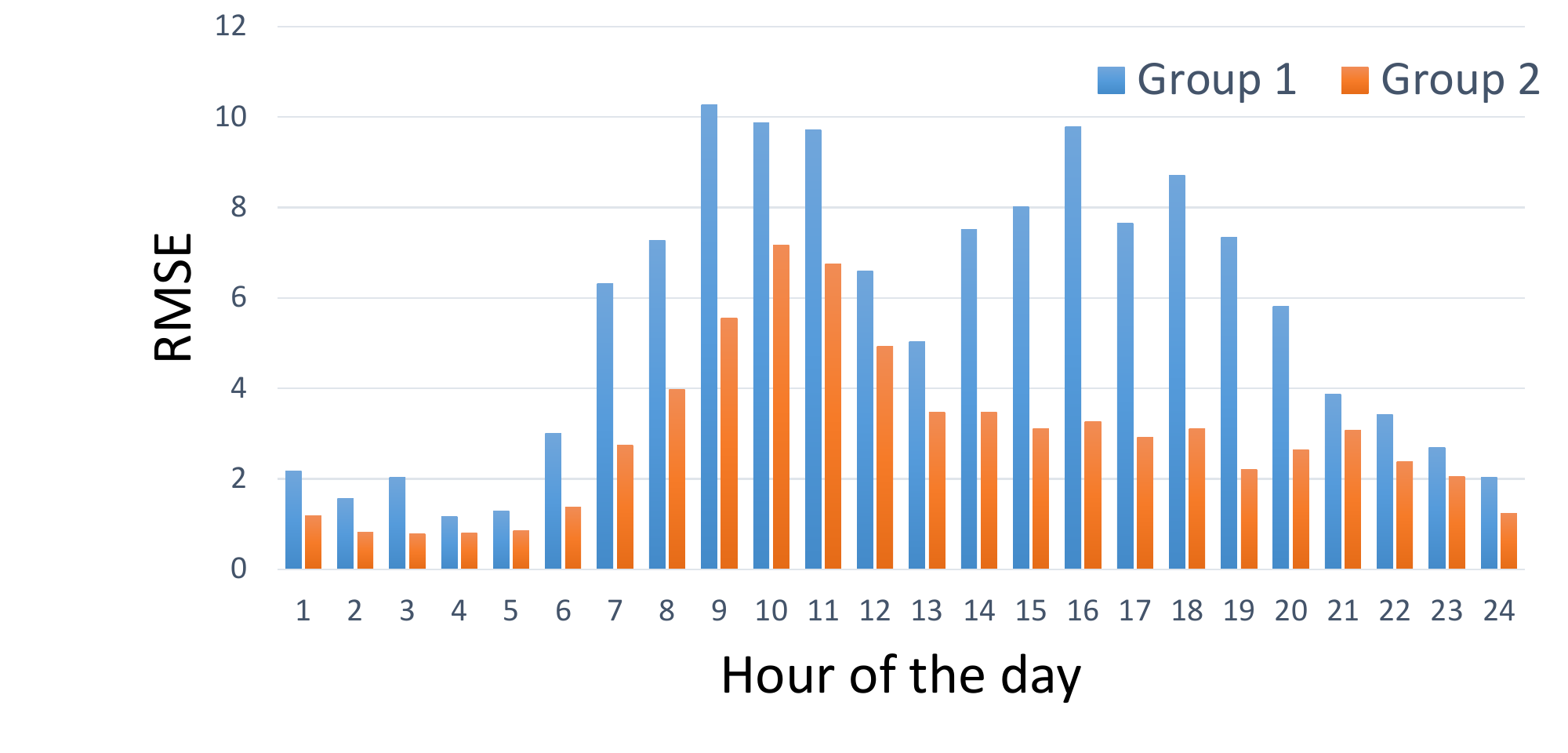}
    \end{subfigure}
    \vskip\baselineskip
    \begin{subfigure}[b]{.45\textwidth}   
        \centering 
        \includegraphics[width=\columnwidth]{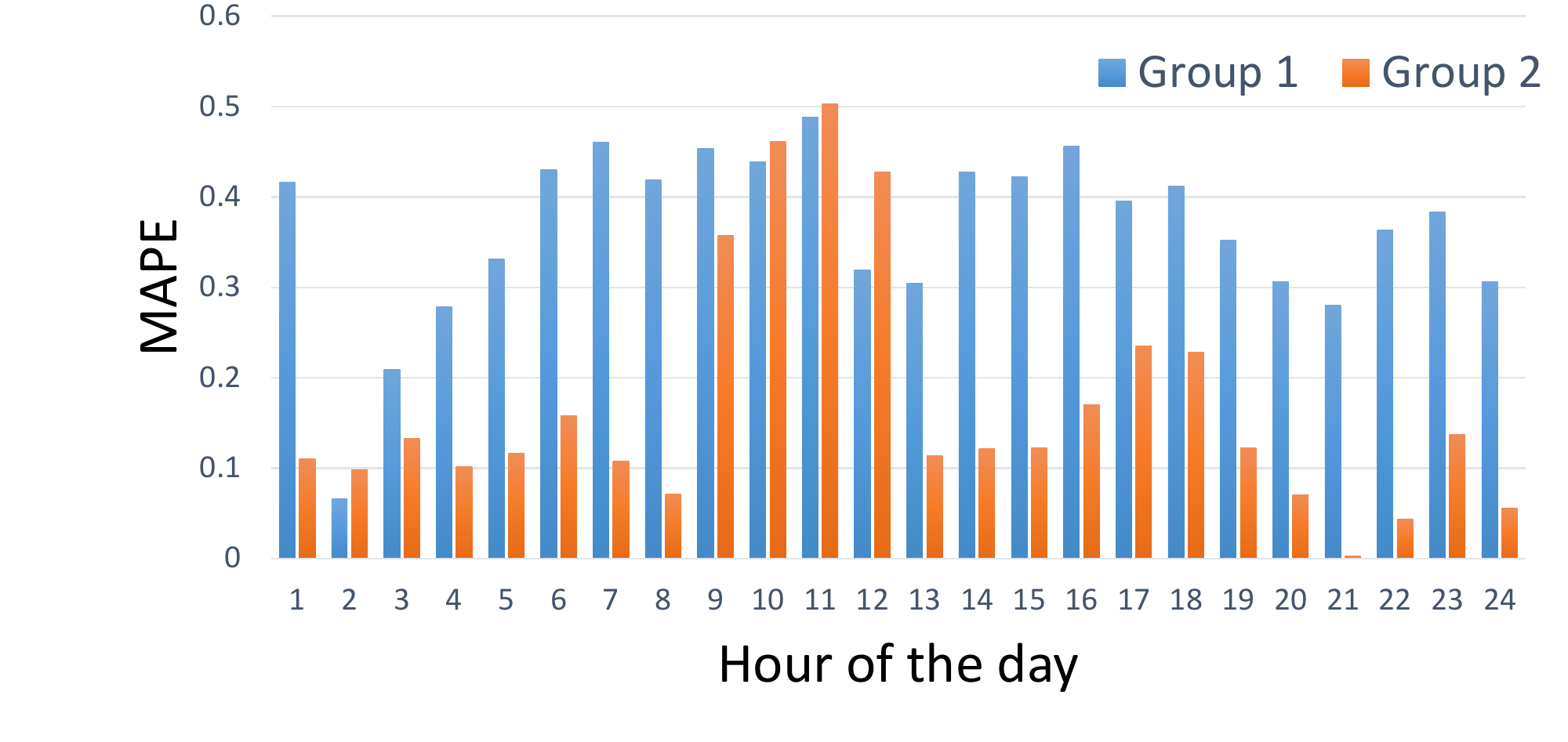}
    \end{subfigure}
    \caption{Error distribution over different hours of a day. Number $i$ on x-axis indicates the $i$-th hour of a day. For example, $7$ means the $7$th hour, i.e., 6:00 to 7:00.}
    \label{temperr}
\end{figure}

Since traffic volume value series are closely related to different times of a day, and peak hours are more interesting, we further explore the performance of \ce at different hours. The averaged RMSE and MAPE values are calculated over each hour of a day in the testing set. Detailed values are illustrated in Figure \ref{temperr}. To better interpret those errors, we still use the same group category in Section \ref{spaterr}. 

As it reads from the figure, the RMSE values for both groups of road segments have clear patterns. The values are higher during the busy time, such as rush hours in the morning and in the afternoon. This is reasonable since common practice tells us that peak hours would have more traffic. And RMSE values are positively correlated to the ground truth values.

On the other hand, the relative error of \ce (i.e., the MAPE values) are pretty stable for Group 1 road segments, where the values of MAPE in rush hours are slightly higher than other ones. This is again due to traffic patterns during rush hours are more uncertain, and more accidental events could change the traffic volume dramatically. Those are reasons why inference difficulty is increased, leading to higher percentage error. At the same time, \ce achieve very low MAPE values for Group 2 road segments at around 0.1 during off-peak hours. And its MAPE values during rush hours in the morning is reasonable. This is explainable since accidental events or traffic jams are less likely to happen near roads from Group 2, because of their less traffic volume.

\section{Conclusion}
In this paper, we propose a novel framework \ce to citywide traffic volume inference using dense and incomplete trajectories. \ce combines high-fidelity traffic simulator with deep reinforcement learning to recover full vehicle movements from incomplete trajectories. 
The masked semi-supervised approach enhanced by multi-view graph embedding is introduced for citywide traffic volume inference.
We evaluate \ce in two regions in a provincial capital city in China to demonstrate its effectiveness.

\bibliographystyle{ACM-Reference-Format}
\balance 
\bibliography{main}


\begin{thebibliography}{00}


\ifx \showCODEN    \undefined \def \showCODEN     #1{\unskip}     \fi
\ifx \showDOI      \undefined \def \showDOI       #1{#1}\fi
\ifx \showISBNx    \undefined \def \showISBNx     #1{\unskip}     \fi
\ifx \showISBNxiii \undefined \def \showISBNxiii  #1{\unskip}     \fi
\ifx \showISSN     \undefined \def \showISSN      #1{\unskip}     \fi
\ifx \showLCCN     \undefined \def \showLCCN      #1{\unskip}     \fi
\ifx \shownote     \undefined \def \shownote      #1{#1}          \fi
\ifx \showarticletitle \undefined \def \showarticletitle #1{#1}   \fi
\ifx \showURL      \undefined \def \showURL       {\relax}        \fi
\providecommand\bibfield[2]{#2}
\providecommand\bibinfo[2]{#2}
\providecommand\natexlab[1]{#1}
\providecommand\showeprint[2][]{arXiv:#2}

\bibitem[\protect\citeauthoryear{Argyriou, Herbster, and Pontil}{Argyriou
  et~al\mbox{.}}{2006}]%
        {argyriou2006combining}
\bibfield{author}{\bibinfo{person}{Andreas Argyriou}, \bibinfo{person}{Mark
  Herbster}, {and} \bibinfo{person}{Massimiliano Pontil}.}
  \bibinfo{year}{2006}\natexlab{}.
\newblock \showarticletitle{Combining graph Laplacians for semi--supervised
  learning}. In \bibinfo{booktitle}{{\em Advances in Neural Information
  Processing Systems}}. \bibinfo{pages}{67--74}.
\newblock


\bibitem[\protect\citeauthoryear{Asif, Mitrovic, Dauwels, and Jaillet}{Asif
  et~al\mbox{.}}{2016}]%
        {asif2016matrix}
\bibfield{author}{\bibinfo{person}{Muhammad~Tayyab Asif},
  \bibinfo{person}{Nikola Mitrovic}, \bibinfo{person}{Justin Dauwels}, {and}
  \bibinfo{person}{Patrick Jaillet}.} \bibinfo{year}{2016}\natexlab{}.
\newblock \showarticletitle{Matrix and tensor based methods for missing data
  estimation in large traffic networks}.
\newblock \bibinfo{journal}{{\em IEEE Transactions on Intelligent
  Transportation Systems\/}} \bibinfo{volume}{17}, \bibinfo{number}{7}
  (\bibinfo{year}{2016}), \bibinfo{pages}{1816--1825}.
\newblock


\bibitem[\protect\citeauthoryear{Aslam, Lim, Pan, and Rus}{Aslam
  et~al\mbox{.}}{2012}]%
        {aslam2012city}
\bibfield{author}{\bibinfo{person}{Javed Aslam}, \bibinfo{person}{Sejoon Lim},
  \bibinfo{person}{Xinghao Pan}, {and} \bibinfo{person}{Daniela Rus}.}
  \bibinfo{year}{2012}\natexlab{}.
\newblock \showarticletitle{City-scale traffic estimation from a roving sensor
  network}. In \bibinfo{booktitle}{{\em Proceedings of the 10th ACM Conference
  on Embedded Network Sensor Systems}}. ACM, \bibinfo{pages}{141--154}.
\newblock


\bibitem[\protect\citeauthoryear{Banerjee, Ranu, and Raghavan}{Banerjee
  et~al\mbox{.}}{2014}]%
        {banerjee2014inferring}
\bibfield{author}{\bibinfo{person}{Prithu Banerjee}, \bibinfo{person}{Sayan
  Ranu}, {and} \bibinfo{person}{Sriram Raghavan}.}
  \bibinfo{year}{2014}\natexlab{}.
\newblock \showarticletitle{Inferring uncertain trajectories from partial
  observations}. In \bibinfo{booktitle}{{\em Data Mining (ICDM), 2014 IEEE
  International Conference on}}. IEEE, \bibinfo{pages}{30--39}.
\newblock


\bibitem[\protect\citeauthoryear{Chen and Guestrin}{Chen and Guestrin}{2016}]%
        {chen2016xgboost}
\bibfield{author}{\bibinfo{person}{Tianqi Chen} {and} \bibinfo{person}{Carlos
  Guestrin}.} \bibinfo{year}{2016}\natexlab{}.
\newblock \showarticletitle{Xgboost: A scalable tree boosting system}. In
  \bibinfo{booktitle}{{\em Proceedings of the 22nd ACM SIGKDD International
  Conference on Knowledge Discovery and Data Mining}}. ACM,
  \bibinfo{pages}{785--794}.
\newblock


\bibitem[\protect\citeauthoryear{Culp and Michailidis}{Culp and
  Michailidis}{2008}]%
        {culp2008graph}
\bibfield{author}{\bibinfo{person}{Mark Culp} {and} \bibinfo{person}{George
  Michailidis}.} \bibinfo{year}{2008}\natexlab{}.
\newblock \showarticletitle{Graph-based semisupervised learning}.
\newblock \bibinfo{journal}{{\em IEEE Transactions on Pattern Analysis and
  Machine Intelligence\/}} \bibinfo{volume}{30}, \bibinfo{number}{1}
  (\bibinfo{year}{2008}), \bibinfo{pages}{174--179}.
\newblock


\bibitem[\protect\citeauthoryear{Duan, Du, Phuoc, and Hoang}{Duan
  et~al\mbox{.}}{2005}]%
        {duan2005building}
\bibfield{author}{\bibinfo{person}{Tran~Duc Duan}, \bibinfo{person}{TL~Hong
  Du}, \bibinfo{person}{Tran~Vinh Phuoc}, {and} \bibinfo{person}{Nguyen~Viet
  Hoang}.} \bibinfo{year}{2005}\natexlab{}.
\newblock \showarticletitle{Building an automatic vehicle license plate
  recognition system}. In \bibinfo{booktitle}{{\em Proc. Int. Conf. Comput.
  Sci. RIVF}}. Citeseer, \bibinfo{pages}{59--63}.
\newblock


\bibitem[\protect\citeauthoryear{G{\"u}hnemann, Sch{\"a}fer, Thiessenhusen, and
  Wagner}{G{\"u}hnemann et~al\mbox{.}}{2004}]%
        {guhnemann2004monitoring}
\bibfield{author}{\bibinfo{person}{Astrid G{\"u}hnemann},
  \bibinfo{person}{Ralf-Peter Sch{\"a}fer}, \bibinfo{person}{Kai-Uwe
  Thiessenhusen}, {and} \bibinfo{person}{Peter Wagner}.}
  \bibinfo{year}{2004}\natexlab{}.
\newblock \showarticletitle{Monitoring traffic and emissions by floating car
  data}.
\newblock  (\bibinfo{year}{2004}).
\newblock


\bibitem[\protect\citeauthoryear{Hu, Li, Bao, Cui, and Feng}{Hu
  et~al\mbox{.}}{2016}]%
        {hu2016crowdsourcing}
\bibfield{author}{\bibinfo{person}{Huiqi Hu}, \bibinfo{person}{Guoliang Li},
  \bibinfo{person}{Zhifeng Bao}, \bibinfo{person}{Yan Cui}, {and}
  \bibinfo{person}{Jianhua Feng}.} \bibinfo{year}{2016}\natexlab{}.
\newblock \showarticletitle{Crowdsourcing-based real-time urban traffic speed
  estimation: From trends to speeds}. In \bibinfo{booktitle}{{\em Data
  Engineering (ICDE), 2016 IEEE 32nd International Conference on}}. IEEE,
  \bibinfo{pages}{883--894}.
\newblock


\bibitem[\protect\citeauthoryear{Kingma and Ba}{Kingma and Ba}{2014}]%
        {kingma2014adam}
\bibfield{author}{\bibinfo{person}{Diederik~P Kingma} {and}
  \bibinfo{person}{Jimmy Ba}.} \bibinfo{year}{2014}\natexlab{}.
\newblock \showarticletitle{Adam: A method for stochastic optimization}.
\newblock \bibinfo{journal}{{\em arXiv preprint arXiv:1412.6980\/}}
  (\bibinfo{year}{2014}).
\newblock


\bibitem[\protect\citeauthoryear{Krajzewicz, Erdmann, Behrisch, and
  Bieker}{Krajzewicz et~al\mbox{.}}{2012}]%
        {sumo2012}
\bibfield{author}{\bibinfo{person}{Daniel Krajzewicz}, \bibinfo{person}{Jakob
  Erdmann}, \bibinfo{person}{Michael Behrisch}, {and} \bibinfo{person}{Laura
  Bieker}.} \bibinfo{year}{2012}\natexlab{}.
\newblock \showarticletitle{Recent Development and Applications of {SUMO -
  Simulation of Urban MObility}}.
\newblock \bibinfo{journal}{{\em International Journal On Advances in Systems
  and Measurements\/}} \bibinfo{volume}{5}, \bibinfo{number}{3\&4}
  (\bibinfo{date}{December} \bibinfo{year}{2012}), \bibinfo{pages}{128--138}.
\newblock


\bibitem[\protect\citeauthoryear{Kwon, Varaiya, and Skabardonis}{Kwon
  et~al\mbox{.}}{2003}]%
        {kwon2003estimation}
\bibfield{author}{\bibinfo{person}{Jaimyoung Kwon}, \bibinfo{person}{Pravin
  Varaiya}, {and} \bibinfo{person}{Alexander Skabardonis}.}
  \bibinfo{year}{2003}\natexlab{}.
\newblock \showarticletitle{Estimation of truck traffic volume from single loop
  detectors with lane-to-lane speed correlation}.
\newblock \bibinfo{journal}{{\em Transportation Research Record: Journal of the
  Transportation Research Board\/}} \bibinfo{number}{1856}
  (\bibinfo{year}{2003}), \bibinfo{pages}{106--117}.
\newblock


\bibitem[\protect\citeauthoryear{Li, Li, and Li}{Li et~al\mbox{.}}{2013}]%
        {li2013efficient}
\bibfield{author}{\bibinfo{person}{Li Li}, \bibinfo{person}{Yuebiao Li}, {and}
  \bibinfo{person}{Zhiheng Li}.} \bibinfo{year}{2013}\natexlab{}.
\newblock \showarticletitle{Efficient missing data imputing for traffic flow by
  considering temporal and spatial dependence}.
\newblock \bibinfo{journal}{{\em Transportation research part C: emerging
  technologies\/}}  \bibinfo{volume}{34} (\bibinfo{year}{2013}),
  \bibinfo{pages}{108--120}.
\newblock


\bibitem[\protect\citeauthoryear{Li, Ahmed, and Smola}{Li
  et~al\mbox{.}}{2015}]%
        {li2015inferring}
\bibfield{author}{\bibinfo{person}{Mu Li}, \bibinfo{person}{Amr Ahmed}, {and}
  \bibinfo{person}{Alexander~J Smola}.} \bibinfo{year}{2015}\natexlab{}.
\newblock \showarticletitle{Inferring movement trajectories from GPS snippets}.
  In \bibinfo{booktitle}{{\em Proceedings of the Eighth ACM International
  Conference on Web Search and Data Mining}}. ACM, \bibinfo{pages}{325--334}.
\newblock


\bibitem[\protect\citeauthoryear{Li, Nie, Wilkie, and Lin}{Li
  et~al\mbox{.}}{2017}]%
        {li2017citywide}
\bibfield{author}{\bibinfo{person}{Weizi Li}, \bibinfo{person}{Dong Nie},
  \bibinfo{person}{David Wilkie}, {and} \bibinfo{person}{Ming~C Lin}.}
  \bibinfo{year}{2017}\natexlab{}.
\newblock \showarticletitle{Citywide estimation of traffic dynamics via sparse
  gps traces}.
\newblock \bibinfo{journal}{{\em IEEE Intelligent Transportation Systems
  Magazine\/}} \bibinfo{volume}{9}, \bibinfo{number}{3} (\bibinfo{year}{2017}),
  \bibinfo{pages}{100--113}.
\newblock


\bibitem[\protect\citeauthoryear{Li, Wang, Yang, and Gong}{Li
  et~al\mbox{.}}{2018}]%
        {Li_2018_ECCV}
\bibfield{author}{\bibinfo{person}{Yandong Li}, \bibinfo{person}{Liqiang Wang},
  \bibinfo{person}{Tianbao Yang}, {and} \bibinfo{person}{Boqing Gong}.}
  \bibinfo{year}{2018}\natexlab{}.
\newblock \showarticletitle{How Local is the Local Diversity? Reinforcing
  Sequential Determinantal Point Processes with Dynamic Ground Sets for
  Supervised Video Summarization}. In \bibinfo{booktitle}{{\em The European
  Conference on Computer Vision (ECCV)}}.
\newblock


\bibitem[\protect\citeauthoryear{Meng, Yi, Su, Gao, and Zheng}{Meng
  et~al\mbox{.}}{2017}]%
        {meng2017city}
\bibfield{author}{\bibinfo{person}{Chuishi Meng}, \bibinfo{person}{Xiuwen Yi},
  \bibinfo{person}{Lu Su}, \bibinfo{person}{Jing Gao}, {and}
  \bibinfo{person}{Yu Zheng}.} \bibinfo{year}{2017}\natexlab{}.
\newblock \showarticletitle{City-wide Traffic Volume Inference with Loop
  Detector Data and Taxi Trajectories}. In \bibinfo{booktitle}{{\em Proceedings
  of ACM International Conference on Advances in Geographical Information
  Systems}}.
\newblock


\bibitem[\protect\citeauthoryear{Mikolov, Chen, Corrado, and Dean}{Mikolov
  et~al\mbox{.}}{2013a}]%
        {mikolov2013efficient}
\bibfield{author}{\bibinfo{person}{Tomas Mikolov}, \bibinfo{person}{Kai Chen},
  \bibinfo{person}{Greg Corrado}, {and} \bibinfo{person}{Jeffrey Dean}.}
  \bibinfo{year}{2013}\natexlab{a}.
\newblock \showarticletitle{Efficient estimation of word representations in
  vector space}.
\newblock \bibinfo{journal}{{\em arXiv preprint arXiv:1301.3781\/}}
  (\bibinfo{year}{2013}).
\newblock


\bibitem[\protect\citeauthoryear{Mikolov, Sutskever, Chen, Corrado, and
  Dean}{Mikolov et~al\mbox{.}}{2013b}]%
        {mikolov2013distributed}
\bibfield{author}{\bibinfo{person}{Tomas Mikolov}, \bibinfo{person}{Ilya
  Sutskever}, \bibinfo{person}{Kai Chen}, \bibinfo{person}{Greg~S Corrado},
  {and} \bibinfo{person}{Jeff Dean}.} \bibinfo{year}{2013}\natexlab{b}.
\newblock \showarticletitle{Distributed representations of words and phrases
  and their compositionality}. In \bibinfo{booktitle}{{\em Advances in neural
  information processing systems}}. \bibinfo{pages}{3111--3119}.
\newblock


\bibitem[\protect\citeauthoryear{Mnih, Kavukcuoglu, Silver, Graves, Antonoglou,
  Wierstra, and Riedmiller}{Mnih et~al\mbox{.}}{2013}]%
        {mnih2013playing}
\bibfield{author}{\bibinfo{person}{Volodymyr Mnih}, \bibinfo{person}{Koray
  Kavukcuoglu}, \bibinfo{person}{David Silver}, \bibinfo{person}{Alex Graves},
  \bibinfo{person}{Ioannis Antonoglou}, \bibinfo{person}{Daan Wierstra}, {and}
  \bibinfo{person}{Martin Riedmiller}.} \bibinfo{year}{2013}\natexlab{}.
\newblock \showarticletitle{Playing atari with deep reinforcement learning}.
\newblock \bibinfo{journal}{{\em arXiv preprint arXiv:1312.5602\/}}
  (\bibinfo{year}{2013}).
\newblock


\bibitem[\protect\citeauthoryear{Mnih, Kavukcuoglu, Silver, Rusu, Veness,
  Bellemare, Graves, Riedmiller, Fidjeland, Ostrovski, et~al\mbox{.}}{Mnih
  et~al\mbox{.}}{2015}]%
        {mnih2015human}
\bibfield{author}{\bibinfo{person}{Volodymyr Mnih}, \bibinfo{person}{Koray
  Kavukcuoglu}, \bibinfo{person}{David Silver}, \bibinfo{person}{Andrei~A
  Rusu}, \bibinfo{person}{Joel Veness}, \bibinfo{person}{Marc~G Bellemare},
  \bibinfo{person}{Alex Graves}, \bibinfo{person}{Martin Riedmiller},
  \bibinfo{person}{Andreas~K Fidjeland}, \bibinfo{person}{Georg Ostrovski},
  {et~al\mbox{.}}} \bibinfo{year}{2015}\natexlab{}.
\newblock \showarticletitle{Human-level control through deep reinforcement
  learning}.
\newblock \bibinfo{journal}{{\em Nature\/}} \bibinfo{volume}{518},
  \bibinfo{number}{7540} (\bibinfo{year}{2015}), \bibinfo{pages}{529}.
\newblock


\bibitem[\protect\citeauthoryear{{OpenStreetMap contributors}}{{OpenStreetMap
  contributors}}{2017}]%
        {OpenStreetMap}
\bibfield{author}{\bibinfo{person}{{OpenStreetMap contributors}}.}
  \bibinfo{year}{2017}\natexlab{}.
\newblock \bibinfo{title}{{OpenStreetMap}}.
\newblock \bibinfo{howpublished}{\url{ https://www.openstreetmap.org }}.
  (\bibinfo{year}{2017}).
\newblock


\bibitem[\protect\citeauthoryear{Qu, Li, Zhang, and Hu}{Qu
  et~al\mbox{.}}{2009}]%
        {qu2009ppca}
\bibfield{author}{\bibinfo{person}{Li Qu}, \bibinfo{person}{Li Li},
  \bibinfo{person}{Yi Zhang}, {and} \bibinfo{person}{Jianming Hu}.}
  \bibinfo{year}{2009}\natexlab{}.
\newblock \showarticletitle{PPCA-based missing data imputation for traffic flow
  volume: A systematical approach}.
\newblock \bibinfo{journal}{{\em IEEE Transactions on intelligent
  transportation systems\/}} \bibinfo{volume}{10}, \bibinfo{number}{3}
  (\bibinfo{year}{2009}), \bibinfo{pages}{512--522}.
\newblock


\bibitem[\protect\citeauthoryear{Qu, Zhang, Hu, Jia, and Li}{Qu
  et~al\mbox{.}}{2008}]%
        {qu2008bpca}
\bibfield{author}{\bibinfo{person}{Li Qu}, \bibinfo{person}{Yi Zhang},
  \bibinfo{person}{Jianming Hu}, \bibinfo{person}{Liyan Jia}, {and}
  \bibinfo{person}{Li Li}.} \bibinfo{year}{2008}\natexlab{}.
\newblock \showarticletitle{A BPCA based missing value imputing method for
  traffic flow volume data}. In \bibinfo{booktitle}{{\em Intelligent Vehicles
  Symposium, 2008 IEEE}}. IEEE, \bibinfo{pages}{985--990}.
\newblock


\bibitem[\protect\citeauthoryear{Qu, Tang, Shang, Ren, Zhang, and Han}{Qu
  et~al\mbox{.}}{2017}]%
        {qu2017attention}
\bibfield{author}{\bibinfo{person}{Meng Qu}, \bibinfo{person}{Jian Tang},
  \bibinfo{person}{Jingbo Shang}, \bibinfo{person}{Xiang Ren},
  \bibinfo{person}{Ming Zhang}, {and} \bibinfo{person}{Jiawei Han}.}
  \bibinfo{year}{2017}\natexlab{}.
\newblock \showarticletitle{An Attention-based Collaboration Framework for
  Multi-View Network Representation Learning}. In \bibinfo{booktitle}{{\em
  Proceedings of the 2017 ACM on Conference on Information and Knowledge
  Management}}. ACM, \bibinfo{pages}{1767--1776}.
\newblock


\bibitem[\protect\citeauthoryear{Ruan, Xu, Sheng, Falkner, Li, and Zhang}{Ruan
  et~al\mbox{.}}{2017}]%
        {ruan2017recovering}
\bibfield{author}{\bibinfo{person}{Wenjie Ruan}, \bibinfo{person}{Peipei Xu},
  \bibinfo{person}{Quan~Z Sheng}, \bibinfo{person}{Nickolas~JG Falkner},
  \bibinfo{person}{Xue Li}, {and} \bibinfo{person}{Wei~Emma Zhang}.}
  \bibinfo{year}{2017}\natexlab{}.
\newblock \showarticletitle{Recovering Missing Values from Corrupted
  Spatio-Temporal Sensory Data via Robust Low-Rank Tensor Completion}. In
  \bibinfo{booktitle}{{\em International Conference on Database Systems for
  Advanced Applications}}. Springer, \bibinfo{pages}{607--622}.
\newblock


\bibitem[\protect\citeauthoryear{Shan, Zhao, and Xia}{Shan
  et~al\mbox{.}}{2013}]%
        {shan2013urban}
\bibfield{author}{\bibinfo{person}{Zhenyu Shan}, \bibinfo{person}{Danna Zhao},
  {and} \bibinfo{person}{Yingjie Xia}.} \bibinfo{year}{2013}\natexlab{}.
\newblock \showarticletitle{Urban road traffic speed estimation for missing
  probe vehicle data based on multiple linear regression model}. In
  \bibinfo{booktitle}{{\em Intelligent Transportation Systems-(ITSC), 2013 16th
  International IEEE Conference on}}. IEEE, \bibinfo{pages}{118--123}.
\newblock


\bibitem[\protect\citeauthoryear{Wang and Li}{Wang and Li}{2017}]%
        {Wang2017}
\bibfield{author}{\bibinfo{person}{Hongjian Wang} {and}
  \bibinfo{person}{Zhenhui Li}.} \bibinfo{year}{2017}\natexlab{}.
\newblock \showarticletitle{{Region Representation Learning via Mobility
  Flow}}.
\newblock   \bibinfo{volume}{10} (\bibinfo{year}{2017}).
\newblock
\showDOI{%
\url{https://doi.org/10.1145/3132847.3133006}}


\bibitem[\protect\citeauthoryear{Wang, Tang, Kuo, Kifer, and Li}{Wang
  et~al\mbox{.}}{2019}]%
        {wang2019simple}
\bibfield{author}{\bibinfo{person}{Hongjian Wang}, \bibinfo{person}{Xianfeng
  Tang}, \bibinfo{person}{Yu-Hsuan Kuo}, \bibinfo{person}{Daniel Kifer}, {and}
  \bibinfo{person}{Zhenhui Li}.} \bibinfo{year}{2019}\natexlab{}.
\newblock \showarticletitle{A simple baseline for travel time estimation using
  large-scale trip data}.
\newblock \bibinfo{journal}{{\em ACM Transactions on Intelligent Systems and
  Technology (TIST)\/}} \bibinfo{volume}{10}, \bibinfo{number}{2}
  (\bibinfo{year}{2019}), \bibinfo{pages}{19}.
\newblock


\bibitem[\protect\citeauthoryear{Wang, Zhang, Gao, Song, Guo, and Shen}{Wang
  et~al\mbox{.}}{2018}]%
        {wang2018mathdqn}
\bibfield{author}{\bibinfo{person}{Lei Wang}, \bibinfo{person}{Dongxiang
  Zhang}, \bibinfo{person}{Lianli Gao}, \bibinfo{person}{Jingkuan Song},
  \bibinfo{person}{Long Guo}, {and} \bibinfo{person}{Heng~Tao Shen}.}
  \bibinfo{year}{2018}\natexlab{}.
\newblock \showarticletitle{MathDQN: Solving Arithmetic Word Problems via Deep
  Reinforcement Learning}.
\newblock  (\bibinfo{year}{2018}).
\newblock


\bibitem[\protect\citeauthoryear{Wang, Zheng, and Xue}{Wang
  et~al\mbox{.}}{2014}]%
        {wang2014travel}
\bibfield{author}{\bibinfo{person}{Yilun Wang}, \bibinfo{person}{Yu Zheng},
  {and} \bibinfo{person}{Yexiang Xue}.} \bibinfo{year}{2014}\natexlab{}.
\newblock \showarticletitle{Travel time estimation of a path using sparse
  trajectories}. In \bibinfo{booktitle}{{\em Proceedings of the 20th ACM SIGKDD
  international conference on Knowledge discovery and data mining}}. ACM,
  \bibinfo{pages}{25--34}.
\newblock


\bibitem[\protect\citeauthoryear{Watkins and Dayan}{Watkins and Dayan}{1992}]%
        {watkins1992q}
\bibfield{author}{\bibinfo{person}{Christopher~JCH Watkins} {and}
  \bibinfo{person}{Peter Dayan}.} \bibinfo{year}{1992}\natexlab{}.
\newblock \showarticletitle{Q-learning}.
\newblock \bibinfo{journal}{{\em Machine learning\/}} \bibinfo{volume}{8},
  \bibinfo{number}{3-4} (\bibinfo{year}{1992}), \bibinfo{pages}{279--292}.
\newblock


\bibitem[\protect\citeauthoryear{Wei, Zheng, Yao, and Li}{Wei
  et~al\mbox{.}}{2018}]%
        {wei2018intellilight}
\bibfield{author}{\bibinfo{person}{Hua Wei}, \bibinfo{person}{Guanjie Zheng},
  \bibinfo{person}{Huaxiu Yao}, {and} \bibinfo{person}{Zhenhui Li}.}
  \bibinfo{year}{2018}\natexlab{}.
\newblock \showarticletitle{Intellilight: A reinforcement learning approach for
  intelligent traffic light control}. In \bibinfo{booktitle}{{\em Proceedings
  of the 24th ACM SIGKDD International Conference on Knowledge Discovery \&
  Data Mining}}. ACM, \bibinfo{pages}{2496--2505}.
\newblock


\bibitem[\protect\citeauthoryear{Wei, Zheng, and Peng}{Wei
  et~al\mbox{.}}{2012}]%
        {wei2012constructing}
\bibfield{author}{\bibinfo{person}{Ling-Yin Wei}, \bibinfo{person}{Yu Zheng},
  {and} \bibinfo{person}{Wen-Chih Peng}.} \bibinfo{year}{2012}\natexlab{}.
\newblock \showarticletitle{Constructing popular routes from uncertain
  trajectories}. In \bibinfo{booktitle}{{\em Proceedings of the 18th ACM SIGKDD
  international conference on Knowledge discovery and data mining}}. ACM,
  \bibinfo{pages}{195--203}.
\newblock


\bibitem[\protect\citeauthoryear{Wilkie, Sewall, and Lin}{Wilkie
  et~al\mbox{.}}{2013}]%
        {wilkie2013flow}
\bibfield{author}{\bibinfo{person}{David Wilkie}, \bibinfo{person}{Jason
  Sewall}, {and} \bibinfo{person}{Ming Lin}.} \bibinfo{year}{2013}\natexlab{}.
\newblock \showarticletitle{Flow reconstruction for data-driven traffic
  animation}.
\newblock \bibinfo{journal}{{\em ACM Transactions on Graphics (TOG)\/}}
  \bibinfo{volume}{32}, \bibinfo{number}{4} (\bibinfo{year}{2013}),
  \bibinfo{pages}{89}.
\newblock


\bibitem[\protect\citeauthoryear{Yamaguchi, Faloutsos, and Kitagawa}{Yamaguchi
  et~al\mbox{.}}{2015}]%
        {yamaguchi2015omni}
\bibfield{author}{\bibinfo{person}{Yuto Yamaguchi}, \bibinfo{person}{Christos
  Faloutsos}, {and} \bibinfo{person}{Hiroyuki Kitagawa}.}
  \bibinfo{year}{2015}\natexlab{}.
\newblock \showarticletitle{OMNI-Prop: Seamless Node Classification on
  Arbitrary Label Correlation.}. In \bibinfo{booktitle}{{\em AAAI}}.
  \bibinfo{pages}{3122--3128}.
\newblock


\bibitem[\protect\citeauthoryear{Yang and Yu}{Yang and Yu}{2016}]%
        {yang2016efficient}
\bibfield{author}{\bibinfo{person}{Ning Yang} {and} \bibinfo{person}{Philip~S
  Yu}.} \bibinfo{year}{2016}\natexlab{}.
\newblock \showarticletitle{Efficient Hidden Trajectory Reconstruction from
  Sparse Data}. In \bibinfo{booktitle}{{\em Proceedings of the 25th ACM
  International on Conference on Information and Knowledge Management}}. ACM,
  \bibinfo{pages}{821--830}.
\newblock


\bibitem[\protect\citeauthoryear{Yao, Tang, Wei, Zheng, and Li}{Yao
  et~al\mbox{.}}{2019}]%
        {yao2019revisiting}
\bibfield{author}{\bibinfo{person}{Huaxiu Yao}, \bibinfo{person}{Xianfeng
  Tang}, \bibinfo{person}{Hua Wei}, \bibinfo{person}{Guanjie Zheng}, {and}
  \bibinfo{person}{Zhenhui Li}.} \bibinfo{year}{2019}\natexlab{}.
\newblock \showarticletitle{Revisiting Spatial-Temporal Similarity: A Deep
  Learning Framework for Traffic Prediction}.
\newblock \bibinfo{journal}{{\em 2019 AAAI Conference on Artificial
  Intelligence (AAAI'19)\/}}.
\newblock


\bibitem[\protect\citeauthoryear{Yao, Wu, Ke, Tang, Jia, Lu, Gong, Ye, and
  Li}{Yao et~al\mbox{.}}{2018}]%
        {yao2018deep}
\bibfield{author}{\bibinfo{person}{Huaxiu Yao}, \bibinfo{person}{Fei Wu},
  \bibinfo{person}{Jintao Ke}, \bibinfo{person}{Xianfeng Tang},
  \bibinfo{person}{Yitian Jia}, \bibinfo{person}{Siyu Lu},
  \bibinfo{person}{Pinghua Gong}, \bibinfo{person}{Jieping Ye}, {and}
  \bibinfo{person}{Zhenhui Li}.} \bibinfo{year}{2018}\natexlab{}.
\newblock \showarticletitle{Deep Multi-View Spatial-Temporal Network for Taxi
  Demand Prediction}.
\newblock \bibinfo{journal}{{\em Proceedings of the Thirty-Second AAAI
  Conference on Artificial Intelligence\/}} (\bibinfo{year}{2018}).
\newblock


\bibitem[\protect\citeauthoryear{Yi, Zheng, Zhang, and Li}{Yi
  et~al\mbox{.}}{2016}]%
        {yi2016st}
\bibfield{author}{\bibinfo{person}{Xiuwen Yi}, \bibinfo{person}{Yu Zheng},
  \bibinfo{person}{Junbo Zhang}, {and} \bibinfo{person}{Tianrui Li}.}
  \bibinfo{year}{2016}\natexlab{}.
\newblock \showarticletitle{ST-MVL: filling missing values in geo-sensory time
  series data}. In \bibinfo{booktitle}{{\em Proceedings of the Twenty-Fifth
  International Joint Conference on Artificial Intelligence}}. AAAI Press,
  \bibinfo{pages}{2704--2710}.
\newblock


\bibitem[\protect\citeauthoryear{Zhan, Li, and Ukkusuri}{Zhan
  et~al\mbox{.}}{2015}]%
        {zhan2015lane}
\bibfield{author}{\bibinfo{person}{Xianyuan Zhan}, \bibinfo{person}{Ruimin Li},
  {and} \bibinfo{person}{Satish~V Ukkusuri}.} \bibinfo{year}{2015}\natexlab{}.
\newblock \showarticletitle{Lane-based real-time queue length estimation using
  license plate recognition data}.
\newblock \bibinfo{journal}{{\em Transportation Research Part C: Emerging
  Technologies\/}}  \bibinfo{volume}{57} (\bibinfo{year}{2015}),
  \bibinfo{pages}{85--102}.
\newblock


\bibitem[\protect\citeauthoryear{Zhan, Zheng, Yi, and Ukkusuri}{Zhan
  et~al\mbox{.}}{2017}]%
        {zhan2017citywide}
\bibfield{author}{\bibinfo{person}{Xianyuan Zhan}, \bibinfo{person}{Yu Zheng},
  \bibinfo{person}{Xiuwen Yi}, {and} \bibinfo{person}{Satish~V Ukkusuri}.}
  \bibinfo{year}{2017}\natexlab{}.
\newblock \showarticletitle{Citywide Traffic Volume Estimation Using Trajectory
  Data}.
\newblock \bibinfo{journal}{{\em IEEE Transactions on Knowledge and Data
  Engineering\/}} \bibinfo{volume}{29}, \bibinfo{number}{2}
  (\bibinfo{year}{2017}), \bibinfo{pages}{272--285}.
\newblock


\bibitem[\protect\citeauthoryear{Zheng, Zhang, Zheng, Xiang, Yuan, Xie, and
  Li}{Zheng et~al\mbox{.}}{2018}]%
        {zheng2018drn}
\bibfield{author}{\bibinfo{person}{Guanjie Zheng}, \bibinfo{person}{Fuzheng
  Zhang}, \bibinfo{person}{Zihan Zheng}, \bibinfo{person}{Yang Xiang},
  \bibinfo{person}{Nicholas~Jing Yuan}, \bibinfo{person}{Xing Xie}, {and}
  \bibinfo{person}{Zhenhui Li}.} \bibinfo{year}{2018}\natexlab{}.
\newblock \showarticletitle{DRN: A Deep Reinforcement Learning Framework for
  News Recommendation}. In \bibinfo{booktitle}{{\em Proceedings of the 2018
  World Wide Web Conference on World Wide Web}}. International World Wide Web
  Conferences Steering Committee, \bibinfo{pages}{167--176}.
\newblock


\bibitem[\protect\citeauthoryear{Zheng and Ni}{Zheng and Ni}{2013}]%
        {zheng2013time}
\bibfield{author}{\bibinfo{person}{Jiangchuan Zheng} {and}
  \bibinfo{person}{Lionel~M Ni}.} \bibinfo{year}{2013}\natexlab{}.
\newblock \showarticletitle{Time-dependent trajectory regression on road
  networks via multi-task learning}. In \bibinfo{booktitle}{{\em Proceedings of
  the Twenty-Seventh AAAI Conference on Artificial Intelligence}}. AAAI Press,
  \bibinfo{pages}{1048--1055}.
\newblock


\bibitem[\protect\citeauthoryear{Zheng, Zheng, Xie, and Zhou}{Zheng
  et~al\mbox{.}}{2012}]%
        {zheng2012reducing}
\bibfield{author}{\bibinfo{person}{Kai Zheng}, \bibinfo{person}{Yu Zheng},
  \bibinfo{person}{Xing Xie}, {and} \bibinfo{person}{Xiaofang Zhou}.}
  \bibinfo{year}{2012}\natexlab{}.
\newblock \showarticletitle{Reducing uncertainty of low-sampling-rate
  trajectories}. In \bibinfo{booktitle}{{\em Data Engineering (ICDE), 2012 IEEE
  28th International Conference on}}. IEEE, \bibinfo{pages}{1144--1155}.
\newblock


\bibitem[\protect\citeauthoryear{Zhou, Bousquet, Lal, Weston, and
  Sch{\"o}lkopf}{Zhou et~al\mbox{.}}{2004}]%
        {zhou2004learning}
\bibfield{author}{\bibinfo{person}{Denny Zhou}, \bibinfo{person}{Olivier
  Bousquet}, \bibinfo{person}{Thomas~N Lal}, \bibinfo{person}{Jason Weston},
  {and} \bibinfo{person}{Bernhard Sch{\"o}lkopf}.}
  \bibinfo{year}{2004}\natexlab{}.
\newblock \showarticletitle{Learning with local and global consistency}. In
  \bibinfo{booktitle}{{\em Advances in neural information processing systems}}.
  \bibinfo{pages}{321--328}.
\newblock


\bibitem[\protect\citeauthoryear{Zhu, Ghahramani, and Lafferty}{Zhu
  et~al\mbox{.}}{2003}]%
        {zhu2003semi}
\bibfield{author}{\bibinfo{person}{Xiaojin Zhu}, \bibinfo{person}{Zoubin
  Ghahramani}, {and} \bibinfo{person}{John~D Lafferty}.}
  \bibinfo{year}{2003}\natexlab{}.
\newblock \showarticletitle{Semi-supervised learning using gaussian fields and
  harmonic functions}. In \bibinfo{booktitle}{{\em Proceedings of the 20th
  International conference on Machine learning (ICML-03)}}.
  \bibinfo{pages}{912--919}.
\newblock


\end{thebibliography}

\end{document}